%% file: main-11231-Lior.tex
\documentclass[11pt,a4paper]{article}
\usepackage{times,latexsym}
\usepackage{url}
\usepackage[T1]{fontenc}

\usepackage[acceptedWithA]{tacl2021v1}
\AddToShipoutPicture{\AtPageLowishCenter{\thepage}}

\input{commands}

\title{Extending Item Response Theory for Efficient and Meaningful Multilingual Evaluation}

\author{
Gili Lior\textsuperscript{1,2}
\quad
Tzviel Frostig\textsuperscript{3}
\quad
Gabriel Stanovsky\textsuperscript{2}
\quad
Matan Eyal\textsuperscript{1}
\\[2mm]
\textsuperscript{1}Google Research
\quad
\textsuperscript{2}The Hebrew University of Jerusalem
\quad
\textsuperscript{3}PhaseV Trials
\\[2mm]
\small{\texttt{gili.lior@mail.huji.ac.il \quad matane@google.com}}
\\
}

\date{}

\begin{document}
\maketitle
\begin{abstract}
\input{sections/00-abstract}
\end{abstract}

\section{Introduction}\label{sec:intro}
\input{sections/01-intro}

\section{Background and Related Work}
\input{sections/02-background}

\section{The \multiirt{} Model}
\label{sec:mirt-framework}
\input{sections/03-multilingual-irt}

\section{Experimental Setup}\label{sec:fit-irt}
\input{sections/04-experimental-setup}

\section{Downstream Applications}\label{sec:applications}
\input{sections/05-practical-implications}

\section{Analysis}\label{sec:properties}
\input{sections/06-properties}

\section{Conclusion}
\input{sections/06-conclusion}

\bibliography{custom}
\bibliographystyle{acl_natbib}

\newpage

\appendix
\input{sections/appendix}

\end{document}

%% file: commands.tex
\usepackage{amsmath}
\usepackage{booktabs}
\usepackage{amssymb}
\usepackage{multirow}
\usepackage{xspace}
\usepackage[framemethod=TikZ]{mdframed}
\usepackage{xcolor}
\usepackage[most]{tcolorbox}
\usepackage{graphicx}
\usepackage{subcaption}
\usepackage{xurl}
\usepackage{tabularx}

\newcommand{\numlangs}{29}
\newcommand{\nummodels}{25}
\newcommand{\numitems}{11{,}759}
\newcommand{\numseeds}{10}
\newcommand{\aboutnumtotal}{${\sim}8.5$M}
\newcommand{\triple}{$(\mathrm{item}, \mathrm{LLM}, \mathrm{language})$}
\newcommand{\nummmlulangs}{29\xspace}
\newcommand{\nummgsmitems}{250\xspace}
\newcommand{\nummgsmlangs}{10\xspace}
\newcommand{\multiirt}{\textit{Multilingual-IRT}}
\newcommand{\parallelirt}{\textit{Parallel-IRT\xspace}}
\newcommand{\indepirt}{\textit{Indep-IRT\xspace}}
\newcommand{\coupledirt}{\textit{Coupled-IRT\xspace}}
\newcommand{\mmluprox}{MMLU-Pro-X\xspace}
\newcommand{\mgsm}{MGSM\xspace}
\newcommand{\gemini}{Gemini-2.5-Flash\xspace}

\newcommand{\aya}{Aya}
\newcommand{\gpt}{GPT-OSS}
\newcommand{\gemma}{Gemma-3}
\newcommand{\phimodel}{Phi-4}
\newcommand{\olmo}{OLMo-3}
\newcommand{\llama}{Llama-3}
\newcommand{\qwen}{Qwen3.5}

\newcommand{\bi}{b_i}
\newcommand{\dil}{d_{il}}
\newcommand{\aione}{a_{i}^{\text{base}}}
\newcommand{\aitwo}{a_{i}^{\text{lang}}}
\newcommand{\airatio}{\aitwo/\aione}

\newcommand{\thetaj}{\theta_j}
\newcommand{\thetajl}{\theta_{jl}}
\newcommand{\epsj}{\boldsymbol{\varepsilon}_j}
\newcommand{\epsjl}{(\varepsilon_j)_l}

\newcommand{\Yijl}{Y_{ijl}}
\newcommand{\pijl}{p_{ijl}}

\newcommand{\Rlng}{R^{\text{lng}}}
\newcommand{\Lcorr}{L_{\text{corr}}}
\newcommand{\siglng}{\sigma}

\newcommand{\logit}{\mathrm{logit}}
\newcommand{\Bernoulli}[1]{\mathrm{Bernoulli}(#1)}

\definecolor{codebackground}{gray}{0.96}

\lstdefinestyle{promptstyle}{
    basicstyle=\scriptsize\ttfamily,
    backgroundcolor=\color{codebackground},
    frame=single,
    rulecolor=\color{codebackground},
    showstringspaces=false,
    breaklines=true,
}

\definecolor{defaccent}{HTML}{2C5282}

\newmdtheoremenv[
  topline=false, bottomline=false, rightline=false, linewidth=1pt,
  linecolor=defaccent,
  backgroundcolor=defaccent!5,
  innertopmargin=6pt,
  innerbottommargin=6pt,
  innerleftmargin=4pt,
  innerrightmargin=4pt,
]{definition}{Definition}

\definecolor{famGermanic}{HTML}{1B9E77}
\definecolor{famRomance}{HTML}{D95F02}
\definecolor{famSlavic}{HTML}{7570B3}
\definecolor{famSouthAsian}{HTML}{E7298A}
\definecolor{famSemitic}{HTML}{66A61E}
\definecolor{famAfrican}{HTML}{E6AB02}
\definecolor{famEastAsian}{HTML}{A6761D}
\definecolor{famSoutheastAsian}{HTML}{1F77B4}
\definecolor{famUralic}{HTML}{B22222}

%% file: sections/00-abstract.tex
Multilingual benchmarks are central to evaluating large language models (LLMs) across languages, but they suffer from three issues: exhaustive evaluation scales linearly with the number of languages, automatic translation introduces errors that are easily missed at scale, and some items conflate general and culture-specific knowledge. We address all three with a unified statistical framework, \multiirt{}, which extends Item Response Theory with per-language difficulty deviations, split discriminability separating content from language effects, and per-language ability residuals. Fitting \multiirt{} on \nummodels{} LLMs across a total of \numlangs{} languages over two multilingual benchmarks covering different domains and task types, we show that the fitted parameters support three practical applications: predicting performance on unobserved instances with up to $16\%$ lower error than the strongest accuracy-based baseline,
identifying translation errors across a wide distribution of languages, and recovering culture-specific items that non-parametric baselines miss.\footnote{Project website~\url{https://gililior.github.io/multilingual_irt/}}

%% file: sections/01-intro.tex
Multilingual benchmarks enable the evaluation of model capabilities across diverse languages, including in low-resource and culture-specific settings~\cite{xu2025survey, qin2025survey}.
To enable direct cross-lingual comparison, such benchmarks are typically parallel, i.e., the same items are tested across multiple languages, so that differences in model performance can be attributed to language- rather than item-variation~\cite{bandarkar-etal-2024-belebele, zhang-etal-2025-p}.

Despite their popularity, current parallel multilingual benchmarks suffer from three fundamental issues. First, the cost of exhaustive evaluation scales linearly with the number of languages in the benchmark, making it prohibitively expensive to evaluate all LLMs on all samples~\cite{wu2025bitter}.
Second, multilingual benchmarks are often built via automatic translation, making them prone to translation errors~\cite{agrawal-etal-2024-translation}, which are hard to identify~\cite{peter2025mind}.
Third,  independently of translation quality, some items conflate between general and culture-specific knowledge, which has been shown to affect model performance~\cite{goldman2025eclektic}.

In this work, we propose a unified framework that jointly reduces evaluation cost, surfaces translation errors, and disentangles general from culture-specific knowledge, by extending Item Response Theory (IRT;~\citealp{lord1952theory}) to multilingual settings.
IRT models the probability that a given test-taker answers a given item correctly, by characterizing each benchmark item~$i$ with a \emph{difficulty} parameter~$b_i$ and a \emph{discriminability} parameter~$a_i$, which captures how strongly the item differentiates between test-takers~\cite{birnbaum1968some}. In parallel, each test-taker~$j$ is parameterized by a \emph{latent ability}~$\theta_j$, and together these parameters define the probability that test-taker $j$ answers item $i$ correctly.

We extend the traditional formulation of IRT to \multiirt{}, with three mathematical components which capture the structure of parallel multilingual benchmarks:
(1) An item's \textit{language-specific difficulty deviations ($d_{il}$)} let its difficulty vary across languages, reflecting how translation can make the same question harder in different languages.
(2) An LLM's (i.e., the test taker) \textit{language-specific ability residual} $\varepsilon_{jl}$ captures how an LLM's proficiency in language $l$ deviates from its overall ability $\theta_j$.
(3) The \textit{content-and-language discriminability split} decomposes an item's discriminability $a_i$ into a language-invariant component $\aione$, capturing how the item's content separates LLMs' performance, and a language-dependent component $\aitwo$, capturing the  separation introduced by the language-specific residual $\varepsilon_{jl}$.
For example, a simple math problem may barely separate LLMs by overall ability (low $\aione$) yet separate them sharply by their proficiency in the given language (high $\aitwo$).
Together, these adaptations assign a probability of correctness to every \triple{} instance.

We fit \multiirt{} on the predictions of \nummodels{} LLMs across \numlangs{} languages on two  multilingual benchmarks: \mmluprox~\cite{xuan-etal-2025-mmlu} , a benchmark for general knowledge multiple-choice questions, and \mgsm{}~\cite{shi2023language}, which consists of open-ended math problems.
We use the fitted parameters to revisit the three issues above, comparing against
non-parametric baselines.

First, \multiirt{} enables efficient evaluation:
across a wide range of subsampling rates, it predicts unseen \triple{} instances with
lower binary cross-entropy ($11{-}16\%$ reduction in \mmluprox, and $3{-}6\%$ in \mgsm) than the strongest accuracy-based baseline, avoiding exhaustive evaluation across all languages while maintaining prediction quality.

Second, ranking items by their per-language difficulty shift $\dil$ surfaces candidate translation errors distributed across all tested languages, unlike accuracy-based baselines, whose detections concentrate in only a few of these languages.

Third, ranking items by the discriminability ratio $\airatio$ recovers culture-specific items that accuracy-based baselines miss.
We also mathematically analyze the fitted parameters, finding, for example, that the discriminability parameters $\aione,\aitwo$ encode signal that simple accuracy-based statistics cannot.

We summarize our contributions:
\begin{itemize}
    \item \multiirt{}, a unified statistical framework for parallel multilingual benchmarks, that jointly models item difficulty, discriminability, and LLM ability across languages.
    \item Three downstream applications addressing key issues in multilingual benchmarks: efficient evaluation, automated flagging of candidate translation errors, and detection of culture-specific items.
    \item An empirical study on \mmluprox{} and \mgsm, covering \numlangs{} languages and \nummodels{} LLMs, showing that IRT-based methods offer advantages over non-parametric accuracy-based baselines.
\end{itemize}

%% file: sections/02-background.tex
In this section we provide the basics of IRT (\S\ref{sec:irt-bg}), and describe how IRT is leveraged for early works in NLP(§\ref{sec:irt-in-nlp}), as well as for more recent LLM benchmarks(\S\ref{sec:irt-in-llm-eval}). Finally, we discuss why standard IRT models are insufficient for multilingual benchmarks (§\ref{sec:irt-limitations}).

\subsection{Item Response Theory (IRT)}\label{sec:irt-bg}

Item Response Theory (IRT) is a statistical framework originally derived from psychometric testing, used to quantify the relationship between an individual's latent ability and their performance on specific test items~\cite{lord1952theory, lord1968statistical}
Under this paradigm, each question in the test
is characterized by item parameters, while each test-taker (in our case, the evaluated LLM) is assigned a latent ability score. In the foundational Two-Parameter Logistic (2PL) IRT model~\cite{birnbaum1968some}, the probability of a correct response for test taker $j$ in question $i$ is:
\begin{equation}\label{eq:irt-base}
p_{ij}=\frac{1}{1+e^{-a_{i}(\theta_{j}-b_{i})}}
\end{equation}
Or equivalently $\logit(p_{ij}) = a_i(\theta_j - b_i)$,
where $\theta_{j}$ is the latent ability of the test taker, and the two item parameters $a_{i}$ and $b_{i}$ represent the discriminability and difficulty of the item, respectively. The difficulty $b_i$ is the ability level at which a test-taker has a $50\%$ chance of success. The discriminability $a_i$ controls how sharply the success probability changes around $b_i$: high $a_i$ means the item cleanly separates test-takers of different abilities, low $a_i$ means it does not.

Building on this, Multidimensional Item Response Theory (MIRT;~\citet{Reckase2009MultidimensionalIR}) extends 2PL IRT by modeling a test-taker's ability as a vector $\boldsymbol{\theta}_{j}$ rather than a scalar, letting an item load onto multiple distinct traits simultaneously:
\begin{equation}\label{eq:mirt}
\logit(p_{ij})=\mathbf{a}_{i}^{\top}\boldsymbol{\theta}_{j}-b_{i}
\end{equation}
where $\mathbf{a}_{i}$ is a vector of discriminability parameters, one per latent trait dimension in $\boldsymbol{\theta}_{j}$.

\leavevmode

\subsection{IRT in NLP}\label{sec:irt-in-nlp}
Early work in NLP adapted IRT to assess human annotator reliability \citep{hovy-etal-2013-learning, otani-etal-2016-irt}. \citet{lalor-etal-2016-building, lalor-etal-2018-understanding, lalor-etal-2019-learning} then formalized treating language models as an ``artificial crowd'' to estimate benchmark difficulty and discriminability, and \citet{rodriguez-etal-2021-evaluation} used IRT parameters to reweight benchmark examples by how well they separate strong from weak models, moving leaderboards beyond raw accuracy. These works established IRT's utility for traditional NLP tasks, but the shift toward LLMs has introduced new scale and efficiency challenges.

\subsection{IRT for efficient LLM evaluation}\label{sec:irt-in-llm-eval}

Recently, IRT has been adopted in NLP primarily to enhance evaluation efficiency. A line of work uses IRT for benchmark compression~\citep{polo2024tinybenchmarks, kipnis2025metabench}, dynamic performance estimation~\citep{hofmann2025fluid}, and adaptive testing to minimize evaluation costs~\citep{li2025adaptive}. Beyond efficiency, IRT is also used to analyze benchmark validity, separating item difficulty from model ability~\citep{zhou2026lost} and testing whether LLM ability and item difficulty interact in the expected way, with stronger models succeeding on harder items, similar to human test-takers~\citep{sauberli-etal-2025-llms, liu2025leveraging}.

These applications, however, have been so far restricted to monolingual (typically English) benchmarks, and to the best of our knowledge has not extended IRT to model the parallel structure of multilingual benchmarks.

\subsection{Limitations of Standard IRT for Multilingual Benchmarks}\label{sec:irt-limitations}
While 2PL MIRT supports multiple latent traits,
it requires non-trivial adaptation for multilingual benchmarks.

First, it suffers from \textit{rotational indeterminacy}: any invertible linear transformation of the latent space yields an equally good fit, so the mapping between latent dimensions and specific traits is not identifiable~\cite{Chen2020EstimationMF, Chen2021ItemRT}. While this can be resolved by specifying in advance which items load on which traits~\cite{Chen2020EstimationMF}, it is unclear how to do so for multilingual benchmarks. This would require to know up front which items probe which language-specific aspects, but this is exactly what to find out using the model.

Second, both standard 2PL IRT and 2PL MIRT do not capture the parallel structure of multilingual benchmarks: translations of the same item are treated as unrelated items, even though answering item $i$ correctly in one language $l$ is correlated with answering it correctly in another language $l'$. These are valuable signals for imputing missing data, detecting translation errors, and identifying culture-specific items, none of which traditional IRT models can exploit.

Our suggested \multiirt{} model addresses both gaps jointly: it resolves rotational indeterminacy through an asymmetric construction of the latent dimensions, and it models the parallel structure of multilingual benchmarks explicitly.

%% file: sections/03-multilingual-irt.tex
In this section, we present our adaptation of the standard IRT model to multilingual benchmarks. We assume a parallel benchmark: each item appears in multiple languages, and the correspondence between an item's versions across languages is known.

\begin{definition}[\multiirt{}]
\label{def:mirt}
For item $i \in \{1, \ldots, I\}$, LLM $j \in \{1, \ldots, J\}$, and language $l \in \{1, \ldots, L\}$, the correctness $Y_{ijl} \in \{0, 1\}$ of LLM $j$ on item $i$ in language $l$ is modeled as $Y_{ijl} \sim \text{Bernoulli}(p_{ijl})$, with
\begin{equation*}
\text{logit}(p_{ijl}) = a_i^{\text{base}}\, \theta_j + a_i^{\text{lang}}\, (\varepsilon_j)_l - (b_i + d_{il}),
\end{equation*}
subject to the sum-to-zero constraints
\begin{equation*}
\sum_{l=1}^{L} d_{il} = 0, \qquad \sum_{l=1}^{L} (\varepsilon_j)_l = 0,
\end{equation*}
with the cross-lingual prior $\varepsilon_j \sim \mathcal{N}(0, \sigma^2 R^{\text{lng}})$, where $R^{\text{lng}} \in \mathbb{R}^{L \times L}$ is a learned correlation matrix.
\end{definition}

\multiirt{} is built around three components, each accommodating important features of parallel multilingual benchmarks not supported by the standard IRT setup.

\leavevmode
\subsection{Three components of \multiirt{}}

\paragraph{(1) Language-specific difficulty deviations $\mathbf{\dil}$.}
Translation can make the same item harder in one language than in another, e.g., when nuance is lost or hints are dropped. \multiirt{} lets item $i$'s difficulty in language $l$ become $b_i + d_{il}$, where $b_i \in \mathbb{R}$ is a baseline difficulty shared across languages and $d_{il}$ captures language-specific deviation. The sum-to-zero constraint on $d_{il}$ makes $b_i$ identifiable as the cross-lingual mean difficulty: $d_{il} > 0$ indicates that translation into language $l$ makes item $i$ harder than its cross-lingual average, and $d_{il} < 0$ that it makes it easier.

\paragraph{(2) Language-specific ability residuals $\mathbf{\varepsilon_j\in \mathbb{R}^L}$ with prior $\mathbf{\Rlng\in\mathbb{R}^{L\times L}}$.}
An LLM's proficiency varies across languages, and prior work has shown that performance is correlated across related languages \citep{ahuja-etal-2023-mega, singh-etal-2025-global}. \multiirt{} decomposes each LLM's per-language ability into a language-invariant overall ability $\theta_j \in \mathbb{R}$ and a language-specific residual vector $\varepsilon_j \in \mathbb{R}^L$, where $(\varepsilon_j)_l$ is LLM $j$'s deviation in language $l$ from its overall ability $\theta_j$. The residuals are coupled across languages through the learned correlation matrix $\Rlng\in\mathbb{R}^{L\times L}$: $\Rlng_{l,l'}$ captures how similarly languages $l$ and $l'$ affect LLM performance, with values close to $1$ indicating that residuals tend in the same direction. $R^{\text{lng}}$ is learned jointly with the rest of the model parameters and provides a data-driven measure of cross-lingual similarity in LLM behavior, without typological or language-family supervision. Beyond its role as a prior, $\Rlng$ itself becomes an interpretable artifact that reveals cross-lingual patterns diverging from established typological groupings.

\paragraph{(3) Decomposed discriminability $\aione,\aitwo$.}
The language of presentation can itself affect how strongly an item discriminates between LLMs: a question solvable by most models in English may separate them sharply once translated into a low-resource language. \multiirt{} decomposes the standard discriminability $a_i$ into two scalars: $\aione$, capturing how much the item's \emph{content} separates LLMs (a language-invariant property), and $\aitwo$, capturing the additional separation introduced by the \emph{change of language}.
In Definition~\ref{def:mirt}, $\aione$
multiplies the overall ability $\theta_j$ and $\aitwo$ multiplies the language-specific residual ($\varepsilon_j$), so the two discriminabilities interact with the corresponding sources of LLMs' variation.

\subsection{Reduced Variants of \multiirt{}}
To quantify the role of each of the three components, we also consider reduced variants of \multiirt{}, including only some of the components.

The variants are summarized in Table~\ref{tab:irt-variants} and given descriptive names for convenience:

\paragraph{\indepirt{}} fits a separate standard IRT model for each language, thereby not modelling cross-lingual relationships.

\paragraph{\parallelirt{}} links corresponding questions across languages through shared item parameters $a_i$ and $b_i$, while modeling language-specific abilities independently through $\theta_{jl}$.

\paragraph{\coupledirt{}} decomposes the language-specific ability into a shared component and a language-specific residual ($\theta_{jl} {\to} \theta_j {+} \varepsilon_{jl}$), with the cross-lingual prior $\Rlng$ placed over the residuals. Unlike the full \multiirt{} model, however, it uses a single discrimination parameter $a_i$ across languages.

\input{tables/variants}

%% file: tables/variants.tex
\begin{table*}[t]
\centering
\resizebox{\textwidth}{!}{%
\begin{tabular}{@{}lcccll@{}}
\toprule
\textbf{Variant Name} & \textbf{Difficulty ($b_i$)} & \textbf{Disc. ($a_i$)} & \textbf{Ability ($\theta_j$)} & \textbf{\# Learned Parameters} & \textbf{Response model ($p_{ijl}$)} \\
\midrule
\indepirt    & $b_{il}$     & $a_{il}$           & $\thetajl$         & $2IL + JL$       & $\sigma\!\left(a_{il}(\thetajl - b_{il})\right)$ \\
\parallelirt & $\bi + \dil$ & $a_i$              & $\thetajl$         & $I(L{+}1) + JL$  & $\sigma\!\left(a_i(\thetajl - (\bi + \dil))\right)$ \\
\coupledirt  & $\bi + \dil$ & $a_i$              & $\thetaj + \epsjl$ & $I(L{+}1) + JL$  & $\sigma\!\left(a_i(\thetaj + \epsjl - (\bi + \dil))\right)$ \\
\midrule
\multiirt    & $\bi + \dil$ & $\aione,\, \aitwo$ & $\thetaj + \epsjl$ & $I(L{+}2) + JL$  & $\sigma\!\left(\aione\thetaj + \aitwo\epsjl - (\bi + \dil)\right)$ \\
\bottomrule
\end{tabular}%
}
\caption{\textbf{The reduced variants of \multiirt}. Each row shows how a variant adapts the parameters of the standard IRT model (Eq.~\ref{eq:irt-base}). All variants are novel to this work. The last row shows the full \multiirt{} model for comparison. \emph{\# Learned parameters} denotes the total number of learned parameters for a given $I$ items, $J$ LLMs, and $L$ languages. Sum-to-zero constraint is applied on $\dil$ and $\epsjl$ wherever they are presented. Moreover, $\epsjl$ is drawn from a shared cross-lingual prior parameterized by $\Rlng$, as presented in Definition~\ref{def:mirt}. Since $\theta_j + \varepsilon_{jl}$ can be absorbed into a single $\theta_{jl}$ during optimization,
\parallelirt{} and \coupledirt{} share the same likelihood. But, they are different
theoretical models: \parallelirt{} treats language abilities as independent,
whereas \coupledirt{} builds each from a shared general ability $\thetaj$ plus a
language-specific residual correlated through $\Rlng$.}
\label{tab:irt-variants}
\end{table*}

%% file: sections/04-experimental-setup.tex
In this section we describe how we fit \multiirt{}
and its reduced variants.

\paragraph{LLMs.} We evaluate \nummodels{} LLMs spanning seven model families: \aya~\citep{ustun-etal-2024-aya}, \qwen~\citep{qwen35blog}, \olmo~\citep{olmo2025olmo}, \gemma~\citep{Kamath2025Gemma3T}, \llama~\citep{grattafiori2024llama}, \phimodel~\citep{abdin2024phi}, and \gpt~\citep{agarwal2025gpt}, ranging in size from 2B to 120B parameters. We use the instruction-tuned variants. Model details are listed in Table~\ref{tab:appendix-models} in the Appendix.

\paragraph{Benchmarks: domains and languages.} We fit \multiirt{} and its
variants on LLM predictions for two popular multilingual benchmarks, covering different domains and task types:
(1)~\mmluprox{}~\citep{xuan-etal-2025-mmlu}, a multiple-choice benchmark of \numitems{} items with 10 answer options per question, spanning diverse domains such as mathematics, physics, and law across \nummmlulangs{} languages; and (2)~\mgsm{}~\citep{shi2023language}, a multilingual grade-school math benchmark of \nummgsmitems{} word problems with free-form numeric answers. It was created via automatic translation of a subset of  GSM8K~\citep{cobbe2021training} from English into \nummgsmlangs{} languages, all of which appear also in \mmluprox{}.
Overall, The \numlangs{} languages in our study include both high-resource (e.g., English, German, Chinese) and low-resource (e.g., Yoruba, Swahili) {} languages, covering a wide range of scripts and families. The full list of languages is given in Table~\ref{tab:appendix-langs} in the Appendix.

\paragraph{Model fit.} We fit the \multiirt{} and its variants using stochastic variational inference~(SVI;~\citet{hoffman2013stochastic}), which approximates the posterior by optimization rather than sampling. This scales to tensors as large as the \aboutnumtotal{} responses of \mmluprox{}, where methods like MCMC would be prohibitively expensive.
Implementation, priors, and optimization details are in Appendix~\ref{sec:reparam}.

\paragraph{Identifiability.} As noted above, \multiirt{} is constructed to break the rotational indeterminacy of standard MIRT models.
We verify this empirically with a simulation study and a seed-stability test (Table~\ref{tab:sim-recovery} and~\ref{tab:seed-stability} in Appendix).
Together these confirm that the \multiirt{} formulation we suggest is not just identifiable in principle but also reproducible in practice.

%% file: sections/05-practical-implications.tex
Below, we show how \multiirt{}'s fitted parameters provide three downstream applications that address core issues with multilingual benchmarks.
We compare \multiirt{} against parametric ablations (our reduced variants) and non-parametric baselines,
namely, accuracy-based scores computed directly from the predictions tensor $\Yijl$ and tailored to each application.

In §\ref{sec:efficient-eval} we find that most  parametric variants provide good predictability over unseen samples, while in §\ref{sec:translation-errors} and §\ref{sec:culture} we show the clear advantage of \multiirt{} over its variants: The full model detects translation errors on a wide range of languages, while the simpler variants detects mostly in low-resource languages, and its unique parametrization allows to detect cultural specific items that are not surfaced by any simpler variant.

\input{figures/partial_obs_two_datasets}

\subsection{Efficient Evaluation}\label{sec:efficient-eval}
We first evaluate how \multiirt{} can help in efficient multilingual benchmarking, where exhaustive evaluation of every LLM on every \triple{} instance of the benchmark can be overly expensive. For instance, \mmluprox{} consists of \numitems{} items in \numlangs{} languages, so evaluating \nummodels{} LLMs against it requires filling an \aboutnumtotal{} predictions tensor $\Yijl$.

\multiirt{} instead can fit its parameters from a partial set of observations and predict the remainder, so most of the tensor never needs to be evaluated. While the model introduces additional parameters to be learned, both estimating them and predicting held-out entries cost negligibly compared to LLM inference (Appendix~\ref{sec:app-compute}).

We sample  $\rho {\in} (0, 1)$ of \triple{} instances on which to run inference, and fit both \multiirt{} and our baseline models on this subset. The learned parameters are then used to predict the success probabilities $\widehat{p}_{ijl}$ on the unobserved $(1 - \rho)$ fraction. We show the results for both \mmluprox{} and \mgsm{}.

\paragraph{\multiirt{} predicts unseen instances better than any non-parametric baseline, mostly due to simple item-parameter sharing of $\mathbf{a_i,b_i}$ across languages.}
As shown in Figure~\ref{fig:partial-obs}, \multiirt{} outperforms every accuracy-based baseline at any fraction of observations $\rho$, for both \mmluprox{} and \mgsm{}. For \mmluprox{} we see an improvement of 11--16\% lower Binary Cross Entropy (BCE) and smaller but consistent gains in ROC-AUC on the unobserved instances. For \mgsm{}, \multiirt{} results in 3.6--5.3\% improvement in BCE.
Examining the  different variants, we see that the biggest difference in performance is between \indepirt{}, which fits each language separately, and \parallelirt{}, which shares item parameters $b_i$ and $a_i$ across languages while allowing per-language difficulty deviations $\dil$. Beyond that, coupling per-language ability residuals through the cross-lingual prior over $\epsjl$ (\coupledirt{}) and splitting discriminability into $\aione, \aitwo$ (\multiirt{}) do not improve prediction substantially.

\input{figures/critical_two_panel}

\paragraph{Predictions are stable in $\mathbf{\rho}$.}
\multiirt{} reaches nearly its full predictive performance from a small fraction of observed instances. BCE improves by only 4.2\%
when going from $\rho{=}0.1$ to $\rho{=}0.9$
of observed instances, with most of the gain achieved by $\rho{=}0.4$. In practice, this means 60\% of the instances can be left unseen with negligible loss in prediction quality.

\leavevmode
\paragraph{Math problems, which do not contain culture information, do not benefit from a language-specific discriminability parameter.}
We see in Figure~\ref{fig:partial-obs}c and d that for \mgsm{}, the variants without the discriminability splitting are performing better than \multiirt{}, especially for the large fraction of observed instances, potentially since to \mgsm{} does not include cultural information, and so the discriminability split parameters are less valuable.

\subsection{Detection of Translation Errors}\label{sec:translation-errors}

Creating multilignual benchmarks via automatic translation may introduce errors that distort the evaluation signal, and the massive scale of such benchmarks makes these artifacts difficult to detect~\cite{peter2025mind}. Because exhaustive manual review is prohibitively expensive, automated methods are needed to flag candidate translation errors for subsequent targeted review by human annotators or LLMs.

\multiirt{}'s lends itself to translation error detection by observing how an instance difficulty diverges from the item's base difficulty $b_i$ via its language-specific difficulty deviations $\dil$. Specifically, errors that alter a question's meaning, remove hints, or invalidate the answer choices would make the item harder. Due to the $\sum_l \dil=0$ constraint, the added difficulty of translation errors may be captured via a positive $\dil$. Based on this, we formulate a ranking score to surface candidate translation errors:
\begin{equation}\label{eq:s-tstat}
s(i, l)  =  \frac{\dil}{\widehat{SE}(\dil)}
\end{equation}
Where $\widehat{SE}(\dil)$ is the posterior standard deviation from SVI. Because this score relies on an LLM-invariant parameter, it captures the translation quality independent of the evaluated LLMs.

\paragraph{LLM-as-a-judge for translation error detection.} We annotate translation errors surfaced by the different methods using \gemini{}~\cite{comanici2025gemini} as a judge  (Full prompt in Listing~\ref{lst:judge-prompt} in Appendix). For each pair of an English source question and its target-language translation, the judge assigns a severity level (Critical, Minor, or None) and selects one of four error categories: Semantic Shift, Logic Alteration, Source Intrusion, or Formatting Failure.
Throughout this section, we use the term \emph{critical-strict} to denote instances labeled as Critical that also fall into one of three categories: Semantic Shift, Logic Alteration, or Formatting Failure. We isolate these three because they fundamentally alter the set of correct answers. Source Intrusion is analyzed separately, since untranslated English terms can make a question easier rather than harder, and do not alter the meaning of the question or its choices as critical errors do.
Figure~\ref{fig:translation-error-examples} in Appendix provides examples and presents the taxonomy of the identified translation errors.

\paragraph{Humans broadly agree with the judge's annotations of translation errors.}
To validate the model's ability to identify translation errors, four in-house annotators evaluated a sample of 77 detected error instances by \gemini{} in \mmluprox{} across Arabic, French, and Russian. Annotators were presented with the judge's full output (error type, severity, and explanation) and were tasked to \textit{agree}, \textit{partially agree}, or \textit{disagree}.
As shown in Table~\ref{tab:translation_validation}, the model demonstrates strong precision across both strict (68\%) and lenient (79\%) agreement criteria, with consistent alignment across error categories, indicating that the judge provides a reliable signal for automated error detection in multilingual benchmarks.
Detailed annotation guidelines are available in Figure~\ref{lst:translation-instructions} in Appendix.

\paragraph{\gemini{}'s translation-error annotations agree with a multi-judge majority}
Following recent work advocating an ensemble approach to LLM-as-a-judge evaluation~\cite{verga2024replacing}, we test whether \gemini{}'s annotations agree with a majority vote over \gemini{}, GPT-5.4~mini~\cite{openai_gpt54mini_2026}, and Claude~Haiku~4.5~\cite{anthropic_haiku45_2025}. We draw a balanced subset of $60$K items annotated by \gemini{}, half flagged with translation errors and half clear, and have the two additional judges annotate them. On this subset, \gemini{} agrees with the majority in 86.3\% of cases.

\paragraph{The errors surfaced by \multiirt{} are largely complementary to those captured by raw accuracy.} Figure~\ref{fig:critical-two-panel} characterizes the top-$1000$ critical-strict errors in \mmluprox{} surfaced by the score $s(i,l)$ (Eq.~\ref{eq:s-tstat}) for \multiirt{} and other  variants, as well as two accuracy-based methods, and a random baseline. While the accuracy methods yield a higher absolute number of critical errors, the pairwise Jaccard overlap between \multiirt{} and the accuracy methods' critical-error sets is notably small (Figure~\ref{fig:critical-two-panel}a for \mmluprox{} and Figure~\ref{fig:jaccard_mgsm} in Appendix for \mgsm{}.).
This indicates that the errors surfaced by \multiirt{} are somewhat orthogonal to those captured by raw accuracy.
The language breakdown (Figure~\ref{fig:critical-two-panel}b) concretely illustrates this divergence: accuracy-based methods concentrate $\geq 70\%$ of their critical error detection in three low-resource African languages (Wolof, Yoruba, and Zulu). In contrast, the \multiirt{} ranking surfaces errors that are distributed much more uniformly across the full $28$-language space, with no single language exceeding $11\%$ of the total and a diverse long tail of languages.

\input{tables/translation-error-judge}

\paragraph{Shared difficulty and decomposed discriminability both extend error detection to higher-resource languages.}
Figure~\ref{fig:critical-two-panel}b traces how the language distribution shifts across the parametric and non-parametric models  for \mmluprox{}.\footnote{This analysis is not relevant to \mgsm{} as all of its \nummgsmlangs{} languages are high resource.} \indepirt{}, which fits each language independently, behaves like the accuracy-based methods, concentrating ${\sim}71\%$ of detections in Wolof, Yoruba, and Zulu. Introducing shared item parameters with per-language deviations $\dil$ (\parallelirt{}) already reduces this concentration to ${\sim}53\%$, surfacing errors in higher-resource languages. Adding the cross-lingual prior $\Rlng$ on top (\coupledirt{}) produces no further change. Only the discriminability decomposition (\multiirt{}) yields a roughly uniform distribution. One possible explanation is that decomposing $a_i$ into $\aione$ and $\aitwo$ frees $\dil$ from absorbing language-dependent discriminability, so that $\dil$ is not inflated for items merely because their language is hard for LLMs in general.

\paragraph{Negative language-specific difficulty deviations identify untranslated English terms in non-English questions.} When $\dil$ is negative, which makes $s(i,l)$ negative also, \multiirt{} signals that a translation is relatively easy compared to its baseline difficulty $b_i$. Interestingly, we find that for \mmluprox{} this negative tail surfaces Source Intrusion errors, i.e., English terms left untranslated in the question. Ranking by the \multiirt{} score $-s(i,l)$, the top $1000$ items contain $16.8\%$ Source Intrusion errors, far more than the best accuracy baseline ($5\%$) or the random baseline ($4\%$). A possible explanation for this difficulty degradation is that untranslated English ``leftovers'' provide some kind of a shortcut for the LLM. Rather than relying on multilingual capabilities, the model leverages its English performance, which inflates its success rate and make the question easier. While raw accuracy struggles to differentiate this artifact from true cross-lingual ability, the \multiirt{} difficulty deviation parameter $\dil$ succeeds.

\subsection{Finding Culture-Specific Items}\label{sec:culture}
Multilingual benchmarks often contain items that conflate general knowledge with culture-specific knowledge. For example, questions about an entity famous only in a particular country, or about historical events tied to a specific nation. Such items have been shown to affect model performance~\cite{goldman2025eclektic}, and flagging them enables filtering or culture-aware evaluation.

\multiirt{}'s discriminability parameters $\aione, \aitwo$ can surface culture-specific items: as defined in \S\ref{sec:mirt-framework}, $\aione$ captures how much an item separates LLMs along a language-invariant axis, while $\aitwo$ captures how much the change of language modulates this separation. We therefore expect culture-specific items to have a large $\aitwo$ relative to $\aione$, which we capture via the ratio $\airatio$.

The analysis below refers only to data from \mmluprox{}, as math questions such as those in \mgsm{} do not contain cultural knowledge and thus such analysis is not applicable.

\paragraph{LLM-as-a-judge for labeling culture specific items.} We use \gemini{} as a judge to annotate the top $2$K culture-specific candidates from each scoring method for \mmluprox{} (full prompt in Listing~\ref{lst:cult-judge-prompt} in the Appendix). The judge sees only the English source instance, and returns whether the item is culture-specific, along with one of six specificity types (Region/Country, Religion/Philosophy, Language-internal, Named-entity, Social-convention, or Universal). Figure~\ref{fig:cultural-specific-examples} in the Appendix provides examples and presents the taxonomy of culture-specific items.

\input{tables/cultural_validation_table}

\paragraph{Human--\gemini{} agreement tracks human confidence.}
To validate the \gemini{} culture-specific labels, seven in-house annotators independently provided 273 annotations (112 unique items). In Table~\ref{tab:cultural_validation}, we report the agreement between the human annotators and the \gemini{} judge. We find that baseline human inter-annotator agreement was only moderate, demonstrating the inherent subjectivity of the task. Overall human--\gemini{} agreement was also moderate, similar to human inter-annotator agreement, but scaled dramatically with human confidence -- when restricting the evaluation to items where annotators reported high confidence, human--\gemini{} agreement rose substantially. Conversely, agreement fell sharply on medium- and low-confidence cases ($n=28$ and $n=5)$. These results suggest \gemini{} is a reliable proxy for unambiguous cultural judgments. Annotation guidelines are available in Figure~\ref{lst:validation-instructions} in Appendix.

\paragraph{The cultural labels are consistent under different judges.}
To test whether the labels reflect the cultural alignment of the particular judge, we re-annotate all items labeled by \gemini{}, with two additional judges from different providers: GPT-5.4~mini~\cite{openai_gpt54mini_2026} and Claude~Haiku~4.5~\cite{anthropic_haiku45_2025}, using an identical prompt and schema. Agreement is high: Fleiss' $\kappa = 0.830$, unanimous on $91.7\%$ of items, with \gemini{} matching the three-judge majority on $97.9\%$.

\paragraph{The decomposed discriminability parameters of \multiirt{} and the accuracy baselines identify largely disjoint sets of culture-specific content.}
As shown in Figure~\ref{fig:cultural-jaccard} in the Appendix, the items surfaced by $\airatio$ barely overlap with those surfaced by either accuracy baseline (Jaccard $\leq 0.11$).
Quantitatively, $\airatio$ surfaces the most culture-specific items ($26.1\%$ in its highest-scored $2$K items), compared to accuracy-based methods ($24.2\%$), with a base rate under random selection of $19.8\%$.

\input{figures/cultural_bi_distribution}

\paragraph{Culture-specific items are harder overall, with a small native-language advantage.} Culture-specific items selected by any of our scoring methods are systematically harder than non-cultural ones (Figure~\ref{fig:cultural-bi-distribution}).
One might further expect culture-specific items to be easier in the language tied to their content, and we find a small advantage on two different tests: (i) U.S./U.K.-tied cultural items are easier in English than non-cultural items are ($+0.035$ vs. $+0.008$, $p < 10^{-5}$); and (ii) items tied to non-English regions ($n=158$) are easier in their own region's language than they would be if each item were randomly reassigned to another region's language ($+0.063$ vs. $+0.037$, $p = 0.028$). The stronger effect for U.S./U.K. items likely reflects the U.S.-centric distribution of \mmluprox{}'s cultural content, as $87.6\%$ of region-tagged cultural items were U.S./U.K.-tied.

%% file: figures/partial_obs_two_datasets.tex
\begin{figure*}[t]
    \centering
    \includegraphics[width=0.78\textwidth]{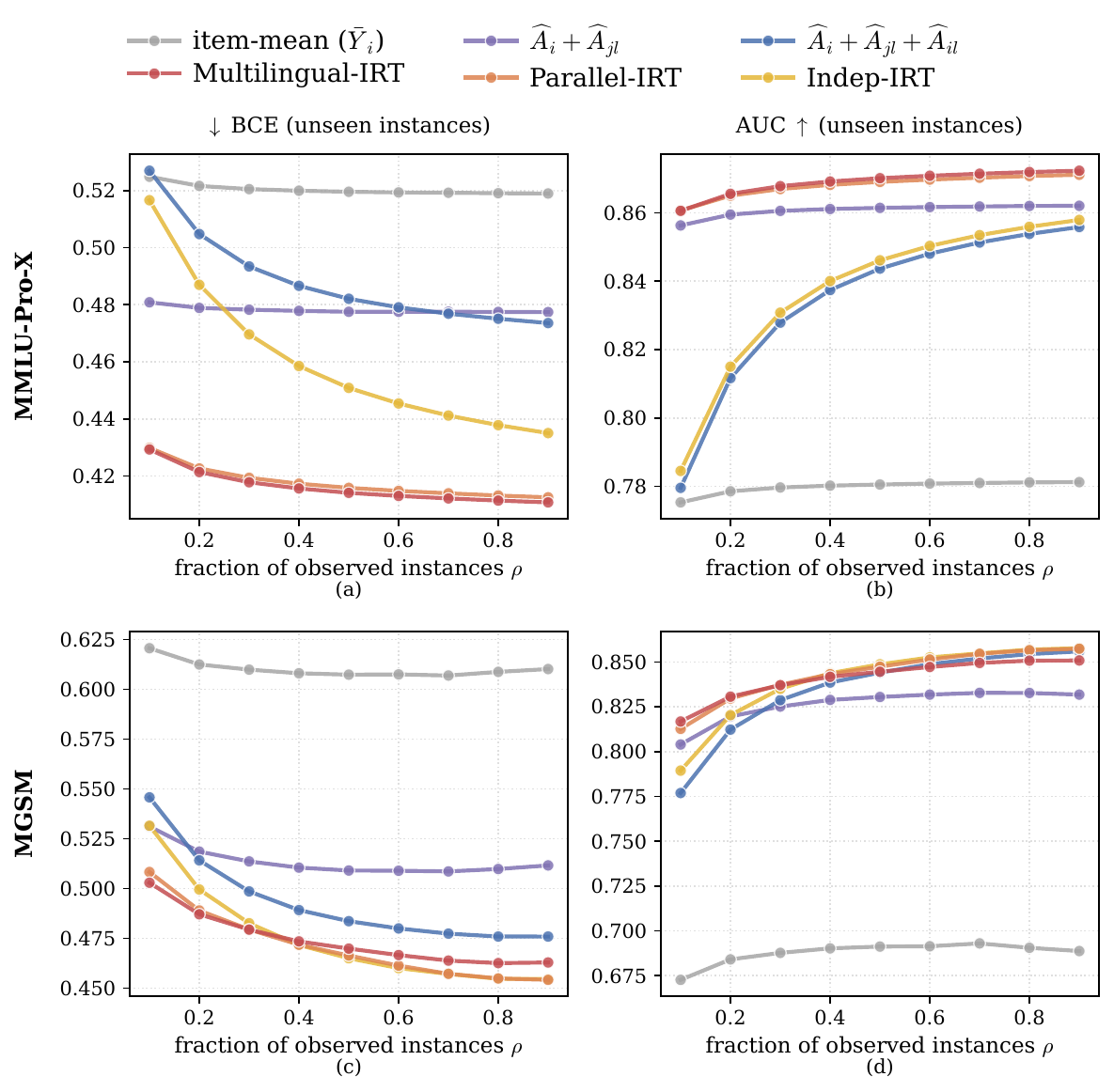}
\caption{\textbf{Prediction scores over unseen \triple{} instances, as a function of the fraction of seen instances.} Two complementary metrics are shown: (a) and (c) are Binary Cross Entropy (BCE), and (b) and (d) are ROC-AUC, where top row is \mmluprox{} and bottom row is \mgsm{}. \coupledirt{}'s curve entirely overlaps with \parallelirt{} and so it is not visible. The accuracy-based non-parametric baselines are computed directly on the predictions tensor $Y_{ijl}$, where $\widehat A_S = \text{logit}(\bar Y_S)$ is the logit of the corresponding mean over the index set $S$, e.g., $\bar Y_i = \text{mean}_{j,l} Y_{ijl}$ (averages over LLMs and languages). The baselines are: (1) per-item mean, $\widehat{p}_{ijl} = \bar Y_i$; (2) logit-additive model, $\widehat{p}_{ijl} = \widehat A_i + \widehat A_{jl}$; and (3) logit-additive model with a per-(item, language) term, $\widehat{p}_{ijl} = \widehat A_i + \widehat A_{jl} + \widehat A_{il}$. }
    \label{fig:partial-obs}
\end{figure*}

%% file: figures/critical_two_panel.tex
\begin{figure*}[t]
    \centering
    \includegraphics[width=\textwidth]{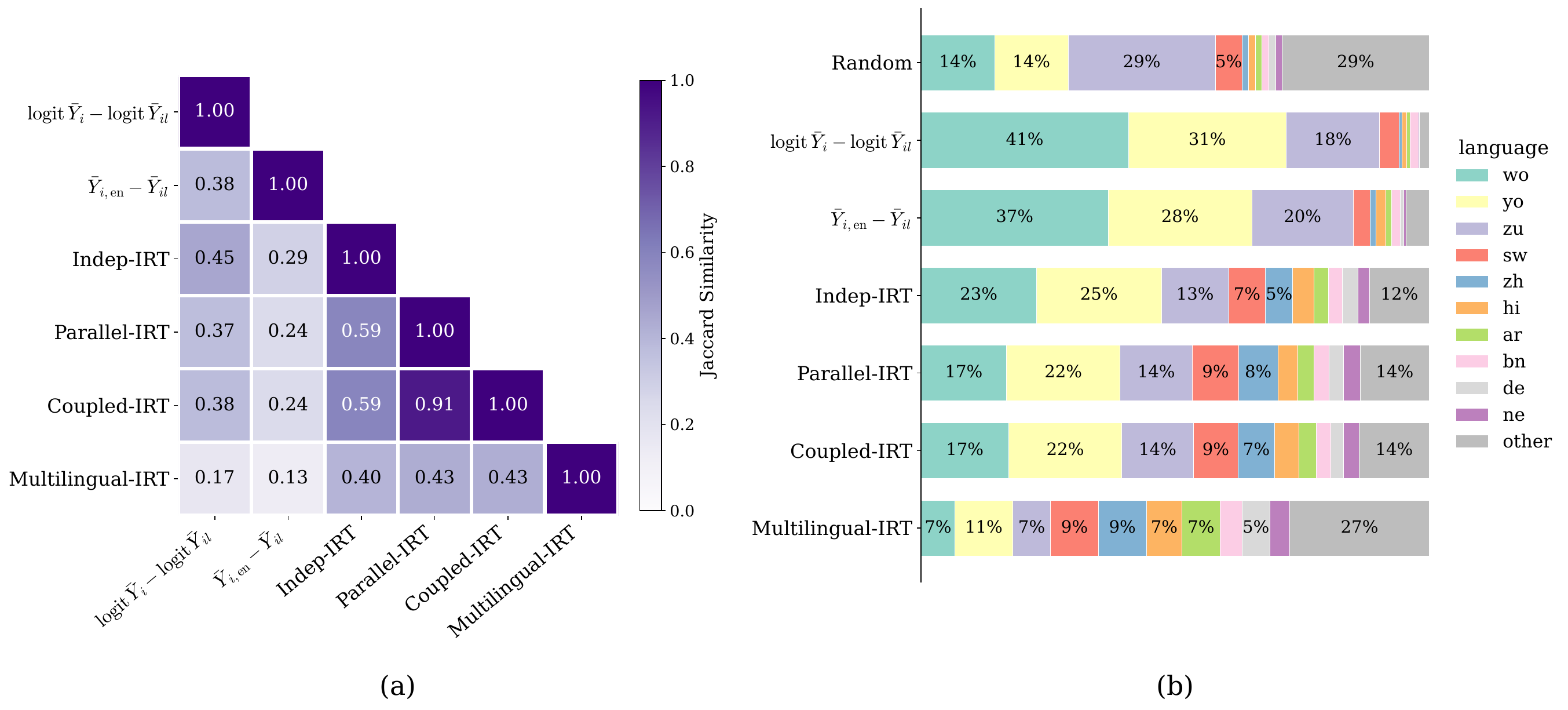}
\caption{\textbf{Top-1000 critical-strict translation errors in \mmluprox{}, surfaced by seven scoring methods  (\S\ref{sec:translation-errors}).} (a) Pairwise Jaccard overlap between top-scored translation errors. (b) Language distribution of errors detected by top-scored items. The IRT difficulty-shift score $s(i,l) = \dil/\widehat{\mathrm{SE}}(\dil)$ is computed for \multiirt{} ($n{=}255$), \coupledirt{} ($n{=}319$), \parallelirt{} ($n{=}309$), and \indepirt{} ($n{=}357$).
The accuracy-based non-parametric baselines are an English-anchored difference $\bar Y_{i,\mathrm{en}}-\bar Y_{il}$ ($n{=}421$) and a logit-additive analog $\logit\,\bar Y_i - \logit\,\bar Y_{il}$ ($n{=}451$). Both metrics are averaged across the $\nummodels$ LLMs, with $\bar Y_{i}$ additionally averaged across all languages. We also include a uniform-random selection over non-English instances as a baseline for the base rate of translation errors ($n{=}76$).}
    \label{fig:critical-two-panel}
\end{figure*}

%% file: tables/translation-error-judge.tex
\begin{table}[t]
    \centering
    \small
    \resizebox{\linewidth}{!}{%
    \begin{tabular}{@{}lccc@{}}
        \toprule
        \textbf{Evaluation Target} & \textbf{Size ($n$)} & \textbf{Strict Prec.} & \textbf{Lenient Prec.} \\
        \midrule
        Overall & 77 & 67.5\% & 79.2\% \\
        \midrule
        \multicolumn{4}{@{}l}{\textit{By Severity}} \\
        Critical & 43 & 55.8\% & 72.1\% \\
        Minor & 34 & 82.4\% & 88.2\% \\
        \midrule
        \multicolumn{4}{@{}l}{\textit{By Error Category}} \\
        Semantic Shift & 51 & 68.6\% & 78.4\% \\
        Logic Alteration & 20 & 60.0\% & 80.0\% \\
        Formatting Failure & 6 & 83.3\% & 83.3\% \\
        \bottomrule
    \end{tabular}%
    }
    \caption{\textbf{Human validation of the \gemini{} translation-error judge.} Results are shown for the overall set of annotations, by severity and error category.}
    \label{tab:translation_validation}
\end{table}

%% file: tables/cultural_validation_table.tex
\begin{table}[t]
    \centering
    \resizebox{\columnwidth}{!}{%
    \begin{tabular}{@{}lccc@{}}
        \toprule
        & \textbf{Human--Human} & \multicolumn{2}{c}{\textbf{Human--\gemini{}}} \\
        \cmidrule(lr){2-2} \cmidrule(l){3-4}
        \textbf{Metric} & (Overlapping) & (Overall) & (High Conf.) \\
        \midrule
        Size ($n$) & 30 & 112 & 79 \\
        Cohen's $\kappa$ & 0.57 (Fleiss') & 0.56 & 0.83 \\
        Raw Agree. & 83\% & 83\% & 95\% \\
        $F_1$ (P / R) & -- & 0.68 (.69/.67) & 0.86 (.86/.86) \\
        \bottomrule
    \end{tabular}%
    }
    \caption{\textbf{Agreement metrics for the cultural specificity task.} The judge's alignment with human annotators scales significantly on high-confidence items. P and R are for precision and recall.}
    \label{tab:cultural_validation}
\end{table}

%% file: figures/cultural_bi_distribution.tex
\begin{figure}[t]
    \centering
    \includegraphics[width=0.8\linewidth]{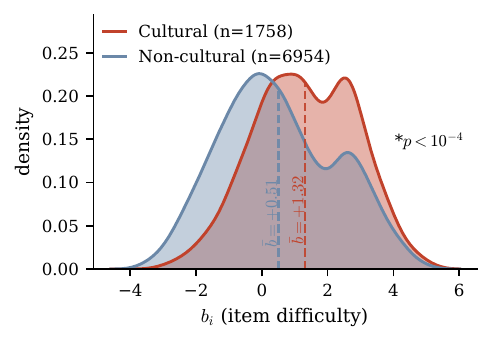}
    \caption{Distribution of $b_i$ for the $n=8712$ items submitted to the \gemini{} judge, split by its cultural-specificity label.}
    \label{fig:cultural-bi-distribution}
\end{figure}

%% file: sections/06-properties.tex
In this section we study what the parameters of \multiirt{} capture, beyond their practical value.
We examine two questions: how do the fitted parameters relate to simple accuracy-based statistics one could compute directly from $\Yijl$ (\S\ref{sec:properties-proxies}), and what cross-lingual structure does the learned correlation matrix $\Rlng$ surface (\S\ref{sec:properties-rlng}).

\subsection{IRT Parameters vs. Accuracy-Based Statistics}\label{sec:properties-proxies}
For each \multiirt{} parameter, we construct a corresponding accuracy-based statistic, by averaging $\Yijl$ along the relevant axis.

Specifically, the mapping is as follows: (1) per-item accuracy for difficulty $\bi$, (2) per-LLM accuracy for ability $\thetaj$, (3) per-(LLM, language) accuracy for the language-specific ability residual $\epsj$, (4) the language-specific deviation in item accuracy from the item's overall mean for $\dil$, and (5) spread statistics ($\text{std}_{j,l}\,\Yijl$ and $\text{std}_l\,\bar Y_{il}$) for the discriminability parameters $\aione$ and $\aitwo$. Full formulas and Pearson correlations against each fitted parameter are in Table~\ref{tab:irt-vs-accuracy}.

\input{tables/correlation_table}

\paragraph{Difficulty and ability parameters match accuracy averages.}
As seen in Table~\ref{tab:irt-vs-accuracy}, the base difficulty and ability parameters $\bi$, $\thetaj$, and $\epsj$ correlate almost perfectly with their accuracy-based matching statistics. For these, the IRT fit largely reproduces what one could compute directly from per-item, per-LLM, and per-(LLM, language) accuracies.
The language-specific difficulty deviation $\dil$ correlates with its empirical statistic more loosely ($r = 0.66$), reflecting a sample-size limitation rather than a gap between IRT and accuracy statistics: a simulation study over larger $J$s shows that $\dil$ recovery climbs from $r{=}0.53$ at $J {=} 25$ to $r {=} 0.89$ by $J {= }240$ (Figure~\ref{fig:identifiability-scaling} in the Appendix).

\paragraph{Discriminability parameters capture signal that accuracy misses.} The discriminability parameters $\aione$ and $\aitwo$ behave differently from the others: their closest accuracy proxies recover only about half of each parameter's variance. They are also unique to \multiirt{} within the reduced variants as no other variant decomposes discriminability, so this signal is inaccessible for both accuracy-based statistics and simpler IRT models.

\subsection{\multiirt{} Recovers Linguistic Structure via $\mathbf{\Rlng}$}\label{sec:properties-rlng}

\input{figures/R_lng_two_datasets}

Beyond modeling, $\Rlng$ provides a data-driven map of cross-lingual interactions in LLM behavior, without any supervision.
Figure~\ref{fig:R-lng} visualizes the fitted correlation matrix $\Rlng$, for both \mmluprox{} and \mgsm{}, {} with languages grouped by family (indicated by colored bounding boxes). Notably, strong positive within-family blocks emerge despite no explicit linguistic family signal being provided during fitting. We observe clear diagonal blocks for Romance (es, fr, it, pt), Slavic (cs, ru, sr, uk), South Asian (bn, hi, mr, ne, te, ur), and African (sw, zu, yo, wo) languages, indicating that the model recovers latent linguistic structure from evaluation data alone. The Indic sub-block within the South Asian family is particularly tight, and the African languages form an exceptionally coherent cluster, despite spanning diverse branches such as Niger-Congo and Bantu.

We also observe that the Germanic family diverges along resource levels: English correlates weakly with other Germanic languages (af, de) and instead clusters with high-resource Semitic (ar) and East Asian languages (ja, zh), which are typologically distant languages that likely share high representation in pre-training corpora. Furthermore, the matrix reveals a notable off-diagonal signal: a strong negative correlation between English and the African languages (wo, yo, zu), suggesting that LLMs which exceed their expected ability in English often underperform in these lower-resource settings. To sum up, $\Rlng$ captures both fine-grained phylogenetic structure and a coarse resource-level gradient, both recovered without supervision.

Moreover, we also observe that $\Rlng$ agrees across benchmarks.
The two fitted $\Rlng$ matrices correlate at $r = 0.70$ over their $55$ shared language pairs (Mantel test, $p < 10^{-4}$), with within-family correlations averaging $0.42$ on \mgsm{} and $0.44$ on  \mmluprox{}. The same cross-lingual structure therefore emerges from a benchmark differing in task format, translation provenance, and scale.

\leavevmode
\paragraph{Stability test under different subsets of models.}
To test whether $\Rlng$ is driven by particular LLMs, and especially by the largest mixture-of-experts (MoE) models, we refit \multiirt{} on model-ablated response tensors of \mmluprox{}, holding the priors, SVI recipe and seed fixed (Table~\ref{tab:rlng-llm-ablation} in Appendix). Since $\Rlng$ moves somewhat even when the data are unchanged, we compare each ablation against three null distributions on the same scale: SVI seed noise, leave-one-LLM-out, and dropping three random LLMs. Dropping the three largest MoE models moves $\Rlng$ no more than dropping any three LLMs does, and the same holds for all four MoE models and for the three highest-ability models. The linguistic structure also survives: the gap between mean within-family and mean between-family correlation is no smaller after any targeted drop than in the full-pool fit. The clustering reported above is therefore robust to the choice of the $\nummodels$ LLMs.
{}

\subsection{Qualitative Analysis}\label{sec:properties-qualitative}

\input{tables/examples}

We now turn from aggregate statistics to individual items, to show what the fitted parameters select for in practice. Table~\ref{tab:qual-examples} samples items from across the fitted parameter grid of \mmluprox{}: nine drawn from the cells of a low/mid/high $\bi$ by low/mid/high $\aione$ grid, and four from the extremes of $\bi$, $\aione$, $\airatio$ and $\mathrm{std}_l\,\dil$. Four more sampled the same way from MGSM, reporting for each its parameters, the percentile each occupies within its own benchmark, the language in which the item is hardest relative to its own difficulty, and whether manual review found a defect.

\paragraph{The top of the difficulty scale includes both hard and unanswerable items.}
Whereas the five lowest-$\bi$ items in \mmluprox{} are straightforward recall or computation questions, three of the eight highest-$\bi$ items are defective: one is incomplete, one is labeled with the wrong answer, and one has no valid answer. Extreme $\bi$ can therefore reflect either genuine difficulty or an unanswerable item.

\paragraph{A low $\aione$ marks items whose answer is contestable.}
As shown in \S\ref{sec:properties-proxies}, low $\aione$ indicates similar success rates across stronger and weaker LLMs. The lowest-$\aione$ items illustrate why: two are labeled with the wrong answer, while another leaves a crucial assumption unspecified. Since $\bi$ and $\aione$ are negatively correlated ($r{=}-0.61$), $1{,}786$ items ($15.2\%$) lie in both the hardest and least-discriminating quartiles, making both parameters useful signals for item inspection.

%% file: tables/correlation_table.tex
\begin{table}[t]
    \centering
    \resizebox{0.9\columnwidth}{!}{%
    \begin{tabular}{lll}
        \toprule
        Parameter & Empirical Statistic & $r$ \\
        \midrule
        $b_i$ & $-\text{logit}\,\bar Y_{i}$ & 0.988 \\
        $d_{il}$ & $-[\text{logit}\,\bar Y_{i l} - \text{logit}\,\bar Y_{i}]$ & 0.664 \\
        $\theta_j$ & $\text{logit}\,\bar Y_{ j}$ & 0.991 \\
        $\varepsilon_{jl}$ & $\text{logit}\,\bar Y_{ jl} - \text{logit}\,\bar Y_{ j}$ & 0.964 \\
        $a_i^{(1)}$ & $\text{std}_{j,l}\,Y_{ijl}$ & 0.548 \\
        $a_i^{(2)}$ & $\text{std}_l\,\overline{Y}_{i l}$ & 0.702 \\
        \bottomrule
    \end{tabular}}
    \caption{Pearson correlation between \multiirt{} learned parameters for \mmluprox and corresponding accuracy-based non-parametric statistics.}
    \label{tab:irt-vs-accuracy}
\end{table}

%% file: figures/R_lng_two_datasets.tex
\begin{figure}[t]
    \centering
    \includegraphics[width=\columnwidth]{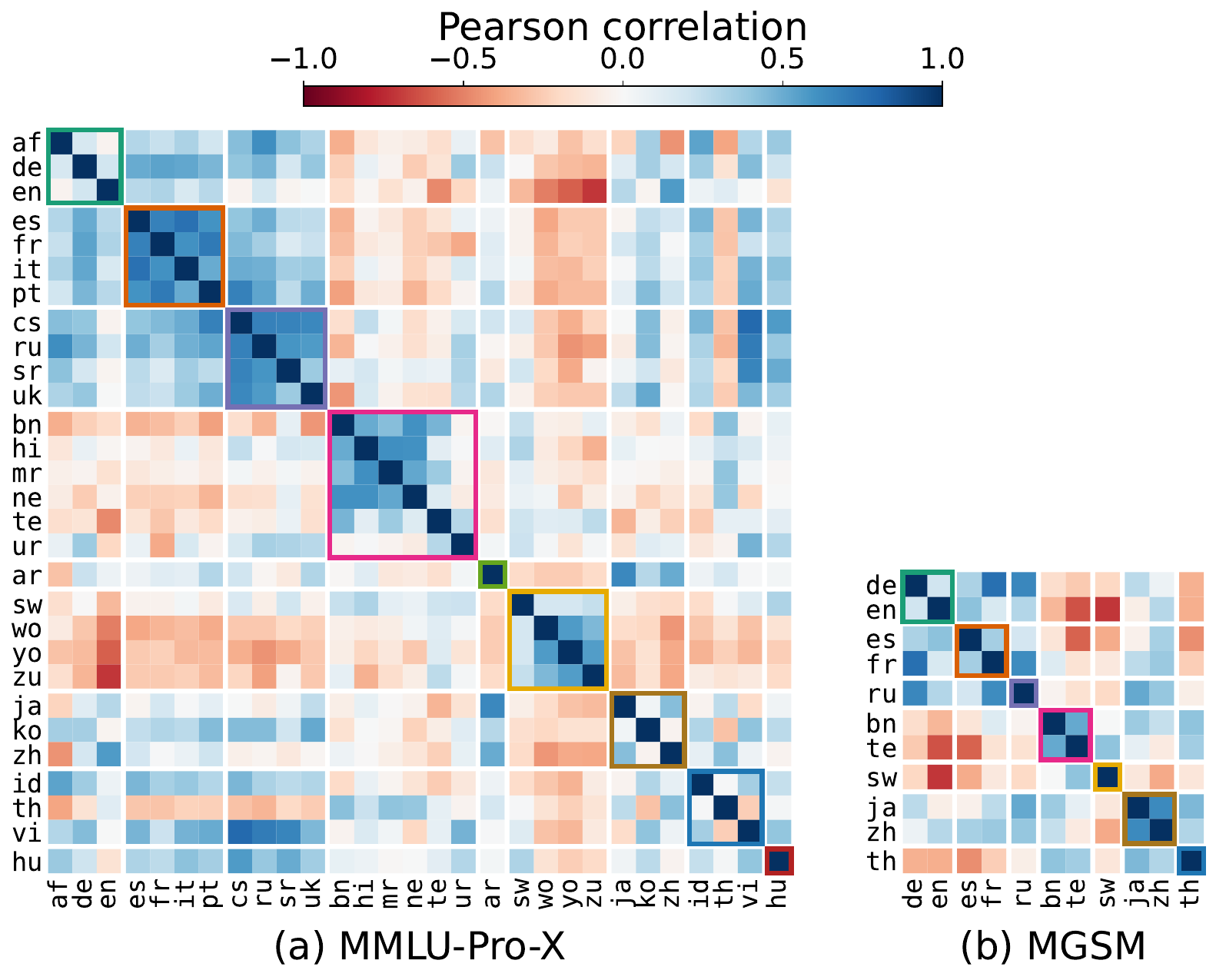}
    \caption{\textbf{Fitted language correlation matrix $\mathbf{\Rlng}$}, the learned prior correlation on per-language ability residuals $\epsj$, for each dataset.
     Colored bounding boxes around the axis labels group languages by linguistic family or geographic region:
     \textcolor{famGermanic}{Germanic}, \textcolor{famRomance}{Romance}, \textcolor{famSlavic}{Slavic}, \textcolor{famSouthAsian}{South Asian}, \textcolor{famSemitic}{Semitic}, \textcolor{famAfrican}{African},  \textcolor{famEastAsian}{ East Asian} , \textcolor{famSoutheastAsian}{Southeast Asian}, \textcolor{famUralic}{Uralic}. No group labels are supplied at fit time.
    }
    \label{fig:R-lng}
\end{figure}

%% file: tables/examples.tex
\begin{table*}[ht!]
    \centering
    \footnotesize
    \resizebox{0.95\textwidth}{!}{%
    \begin{tabular}{@{}lrrrcc@{}}
        \toprule
        \textbf{Item, quoted in English} & $\bi$ & $\aione$ & $\aitwo$ & \textbf{hardest $\dil$} & \textbf{Erroneous} \\
        \midrule
        \multicolumn{6}{@{}l}{\textit{Low $\bi$}} \\
        ``Add. 46,911 + 653,092''                   & $-1.03$ {\tiny(16)} & $0.44$ {\tiny(9)}  & $0.51$ {\tiny(5)}  & ar $+0.37$ & no  \\
        ``$(1+i)^{10} =$''                          & $-1.28$ {\tiny(13)} & $1.02$ {\tiny(44)} & $0.65$ {\tiny(17)} & ru $+0.27$ & no  \\
        ``At a Nash equilibrium,''                  & $-1.66$ {\tiny(8)}  & $1.91$ {\tiny(94)} & $2.01$ {\tiny(95)} & te $+0.36$ & no  \\
        \midrule
        \multicolumn{6}{@{}l}{\textit{Mid $\bi$}} \\
        ``Define poverty.''                         & $+0.67$ {\tiny(51)} & $0.52$ {\tiny(14)} & $0.69$ {\tiny(22)} & it $+0.40$ & yes \\
        ``Homo erectus was:''                       & $+0.55$ {\tiny(48)} & $1.17$ {\tiny(53)} & $0.54$ {\tiny(7)}  & zh $+0.44$ & no  \\
        ``Castration in adulthood''                 & $+0.82$ {\tiny(54)} & $1.79$ {\tiny(89)} & $1.29$ {\tiny(70)} & sw $+0.35$ & no  \\
        \midrule
        \multicolumn{6}{@{}l}{\textit{High $\bi$}} \\
        ``An ecclesia is:''                         & $+2.72$ {\tiny(89)} & $0.43$ {\tiny(8)}  & $0.55$ {\tiny(8)}  & ja $+0.22$ & no  \\
        ``Bronchial breathing is characterized by'' & $+2.95$ {\tiny(91)} & $1.06$ {\tiny(47)} & $0.82$ {\tiny(36)} & ja $+0.16$ & no  \\
        ``The Egyptian system of hieroglyphics:''   & $+2.83$ {\tiny(90)} & $1.71$ {\tiny(85)} & $1.14$ {\tiny(61)} & es $+0.17$ & no  \\
        \midrule
        \multicolumn{6}{@{}l}{\textit{Parameter extremes}} \\
        ``Objects that absorb light appear'' {\tiny(max $\airatio$)}
                                                    & $+0.91$ {\tiny(56)} & $0.40$ {\tiny(6)}  & $1.28$ {\tiny(69)} & ja $+0.36$ & yes \\
        ``Taphonomy is:'' {\tiny(max $\mathrm{std}_l\,\dil$)}
                                                    & $+0.12$ {\tiny(38)} & $1.56$ {\tiny(77)} & $1.51$ {\tiny(81)} & ko $+0.76$ & no  \\
        ``\dots{} increasing $\lambda$ \underline{\hspace{1em}} the testing error'' {\tiny(max $\bi$)}
                                                    & $+5.24$ {\tiny(100)}& $0.82$ {\tiny(33)} & $0.78$ {\tiny(32)} & en $+0.06$ & yes \\
        ``Express (2/5) and [1/(100)] as a percentage'' {\tiny(min $\aione$)}
                                                    & $+1.27$ {\tiny(64)} & $0.15$ {\tiny(0)}  & $0.41$ {\tiny(1)}  & pt $+0.22$ & yes \\
        \bottomrule
    \end{tabular}%
    }
    \caption{\textbf{\mmluprox{} items sampled across the fitted parameter grid.}
The first nine rows come from a low/mid/high $\bi$ $\times$ low/mid/high $\aione$
grid, with rows are ordered by
increasing $\aione$. The last four rows are extremes of $\airatio$,
$\mathrm{std}_l\,\dil$, $\bi$ and $\aione$, as marked. Parenthesized numbers are
percentile ranks among the \numitems{} items, e.g.\ $\bi = -1.03$ {\tiny(16)} is the
$16$th percentile of difficulty. ``Hardest $\dil$'' is the language with the largest
positive difficulty deviation; ``Erroneous'' flags items our manual review of the stem,
options and key found to have multiple correct answers or none.}
    \label{tab:qual-examples}
\end{table*}

%% file: sections/06-conclusion.tex
We introduced \multiirt, an Item Response Theory extension for parallel multilingual benchmarks that models per-language difficulty deviations $\dil$, content and language discriminability ($\aione$, $\aitwo$), and correlated per-language ability residuals $\epsj$ via $\Rlng$. Across \nummodels{} LLMs and \numlangs{} languages of \mmluprox~and \mgsm, \multiirt{} improves prediction, surfaces translation errors, and identifies culture-specific items, offering a practical foundation for multilingual benchmark auditing and use.

\paragraph{Future work} could extend \multiirt{} to evaluate continuous generative scores. This would require adapting the observation model and the corresponding loss function used for fitting, while our core multilingual latent structure would remain entirely unchanged.

\paragraph{Limitations.}
\multiirt{} requires parallel items across languages, so it does not extend to benchmarks constructed independently per language. It is also sample-size sensitive: with \nummodels{} LLMs, $\dil$ and the off-diagonal entries of $\Rlng$ are less precisely recovered and would benefit from larger model pools, though they are effective as is.

\section*{Acknowledgments}
This work was conducted during the first author's (Gili Lior) internship at Google Research, with Matan Eyal as the internship host. We would like to thank the colleagues at Mundo team in Google Research for helpful discussions and feedback throughout this work -- Guy Mor-Lan, Reut Tsarfaty, Dan Malkin, Matan Danos, Idan Brusilovsky, Uri Shaham, Arie Cattan, Adi Mayrav Gilady, and Sivan Eiger. This work was also partially supported by a Google Research Gift to the Hebrew University of Jerusalem and by the Israel Science Foundation (ISF) grant no. 1023/26.

%% file: sections/appendix.tex
\clearpage

\section{Models and Languages}

\input{tables/models_table}

\input{tables/languages_table}

\section{Hyperparameters and Priors}\label{sec:reparam}

This appendix gives the full prior specification and optimization recipe used for every \multiirt{} fit reported in the paper. All models are implemented in NumPyro~\citep{phan2019composable}.

\paragraph{Priors.} We use the following weakly-informative priors. Item discriminabilities are log-normal:
\[
\log \aione,\ \log \aitwo \sim \mathcal{N}(0,\, \sigma_{\log a}^2), \  \sigma_{\log a} = 0.35,
\]
so that each $a_i^{(d)}$ stays positive and is shrunk toward $1$. Baseline item difficulty and per-language difficulty residuals are:
\[
\begin{array}{l}
\bi \sim \mathcal{N}(0,\, \sigma_b^2),\quad \sigma_b = 1.0, \\
\tilde \dil \sim \mathcal{N}(0,\, \sigma_d^2),\quad \sigma_d = 0.30,
\end{array}
\]
with $\tilde \dil$ projected to satisfy the sum-to-zero constraint of Definition~\ref{def:mirt} (see reparameterization below). The model's overall ability follows:
\[
\thetaj \sim \mathcal{N}(0,\, \sigma_\mu^2),\qquad \sigma_\mu = 0.5.
\]
The cross-language ability residuals $\epsj \in \mathbb{R}^L$ are multivariate normal,
\[
\begin{array}{l}
\epsj \sim \mathcal{N}\!\left(\mathbf{0},\, \siglng^2\, \Rlng\right), \\
\siglng \sim \mathrm{HalfNormal}(0.5),
\qquad
\Rlng \sim \mathrm{LKJ}(2),
\end{array}
\]
with $\epsj$ projected to sum-to-zero across $l$. The $\mathrm{LKJ}(2)$ concentration biases $\Rlng$ mildly toward the identity, regularizing the $\binom{L}{2}{=}\binom{\numlangs}{2}{=}406$ off-diagonal entries that must be estimated from $J{=}\nummodels$ per-LLM residual vectors.

\paragraph{Reparameterizations.} Two standard reparameterizations make the posterior easier to fit without changing the model semantically.

\emph{(i) Sum-to-zero by mean-centering.} The constraints on $\dil$ and $\epsjl$ (Definition~\ref{def:mirt}) are enforced by sampling unconstrained variables and subtracting their cross-language mean: e.g.\ $\tilde \dil \stackrel{\mathrm{iid}}{\sim} \mathcal{N}(0,\sigma_d^2)$ and
\[
\dil = \tilde \dil - \tfrac{1}{L}\sum_{l'=1}^L \tilde d_{il'},
\]
and analogously for $\epsj$. This is a hard constraint imposed by construction (not a soft penalty) and -- in contrast to the alternative $(L{-}1)$-free + negated-sum form -- gives every language the same marginal variance $\sigma_d^2(L{-}1)/L$, avoiding the likelihood saturation at large $L$ that the asymmetric form induces when the language singled out as ``last'' accumulates variance $(L{-}1)\sigma_d^2$.

\emph{(ii) Non-centered MVN for $\epsj$.} Let $\Lcorr$ be the lower Cholesky factor of $\Rlng$. With standardized residuals $z_j \sim \mathcal{N}(\mathbf{0}, I_L)$ we sample
\[
\begin{array}{l}
\epsj^{\text{raw}} = \siglng \cdot \Lcorr\, z_j, \\
\epsjl = (\epsj^{\text{raw}})_l - \tfrac{1}{L}\sum_{l'} (\epsj^{\text{raw}})_{l'},
\end{array}
\]
which has the same marginal distribution but couples the latents through a fixed linear transform rather than through a covariance, removing the funnel pathology that plain centered MVN priors induce in SVI. We accordingly place the prior on the Cholesky factor directly, $\Lcorr \sim \mathrm{LKJCholesky}(L, 2)$, which is equivalent to $\Rlng \sim \mathrm{LKJ}(2)$.

\paragraph{Optimization.} We optimize the evidence lower bound (ELBO) with the Adam optimizer~\citep{kingma2014adam} at learning rate $10^{-2}$ for $30{,}000$ steps. The variational family is a rank-$10$ multivariate normal over all latent variables (\texttt{AutoLowRankMultivariateNormal} in NumPyro). The guide is initialized at the prior median (\texttt{init\_to\_median}); we found that the default uniform initialization (\texttt{init\_to\_uniform} with radius $2$) saturated the Bernoulli logits at $L{=}\numlangs{}$ and prevented convergence, because the initial $a_i \cdot \thetajl$ products land in the tens, pushing $\logit(\pijl)$ to $\pm 30$ and the per-instance log-likelihood to $-\infty$. After fitting, we draw $1{,}000$ samples from the guide and report the per-site posterior mean and standard deviation, including deterministic sites ($\dil$, $\epsjl$, $\Rlng = \Lcorr\Lcorr^\top$). Every \multiirt{} fit reported in \S\ref{sec:fit-irt} uses the same SVI seed; the seed-stability check in Appendix~\ref{sec:appendix-identifiability} sweeps $S{=}10$ alternative seeds with the rest of this recipe held fixed.

\section{LLM-as-a-Judge}\label{sec:app-llm-aaj}
Both the translation-error audit (\S\ref{sec:translation-errors}) and the cultural-specificity audit (\S\ref{sec:culture}) use \gemini{} via the Gemini Batch API. We chose \gemini{} for its multilingual coverage (the translation-error judge sees source-target pairs across all $28$ non-English languages), its low cost (${\sim}\$24$ total for both audits combined; see Appendix~\ref{sec:app-compute}), and its support for structured JSON output via the Gemini \texttt{response\_schema} parameter, which constrains each judgment to a fixed schema and eliminates parsing errors. All judgments use the default temperature setting from the Gemini Batch API. We run a single judge pass per instance and do not measure intra-judge consistency. The judge prompts  are shown below (Listings~\ref{lst:judge-prompt} and \ref{lst:cult-judge-prompt}).

\section{Compute Cost}\label{sec:app-compute}
We document the compute used across three categories: open-source LLM inference, IRT model fitting, and closed-source API judging.

\paragraph{LLM Inference.}\label{sec:app-compute-inference} We evaluated \nummodels{} open-source LLMs on \mmluprox{} across \numlangs{} languages ($\numitems{}$ items per language), producing \aboutnumtotal{} item-level predictions, using the LM Evaluation Harness~\cite{eval-harness} with a vLLM backend~\cite{kwon2023efficient}. GPU allocation followed model size: models up to $\sim$11B (both Aya 3B/8B, Qwen3.5-2B/4B, Llama-3.1-8B/3.2-3B, Phi-4-mini instruct and reasoning, gemma-3-4b/12b, Olmo-3-7B instruct/think, Phi-4-reasoning, Phi-4-reasoning-plus, Qwen3.5-9B) ran on a single GPU (24--48~GB); mid-size models (Olmo-3.1-32B instruct/think, gemma-3-27b, gpt-oss-20b, Qwen3.5-27B/35B-A3B, aya-expanse-32b) on 2 GPUs; and the three largest (gpt-oss-120b, Llama-3.3-70B, Qwen3.5-122B-A10B) on 8 A100 80~GB GPUs.

\paragraph{IRT Model Fitting.}\label{sec:app-compute-fitting} All IRT fits use NumPyro stochastic variational inference (SVI) on a single NVIDIA RTX A5000 (24~GB), with a low-rank Gaussian guide (rank 10), learning rate $10^{-2}$, and 30{,}000 SVI steps unless noted; wall-clock equals GPU-hours since each fit uses one GPU. The full set of fits totals 366 runs and $\sim$7.7 GPU-hours, dominated by the partial-observation sweep (270 fits, $\sim$6.0~h) and the identifiability-scaling study (20 fits, 0.95~h); the main \multiirt{}, \parallelirt{}, and \indepirt{} fits together account for under 0.1~h.

\paragraph{Closed-Source API Judging.}\label{sec:app-compute-api} The cultural-specificity (\S\ref{sec:culture}) and translation-error (\S\ref{sec:translation-errors}) judges use Gemini-2.5-Flash via the Gemini Batch API ($\sim$285K calls, $\sim$225M input and $\sim$23M output tokens total). The translation-error judge dominates (275K calls, 220M/22M tokens) over the cultural-specificity judge (8.7K calls, 4.4M/0.7M tokens). At Batch API rates (\$0.075 and \$0.30 per 1M input/output tokens), the estimated cost is \$23.20 and \$0.54 respectively, $\sim$\$23.80 total.

\newpage
\input{figures/judge_prompt}

\input{figures/cultural_judge_prompt}

\newpage

\input{figures/translation_instructions}

\input{figures/validation_instructions}

\clearpage

\section{Complementary Results}

\input{figures/translation_overlap_mgsm}

\input{figures/cultural_jaccard}

\input{tables/stability_test_for_subsets}

\section{Empirical Identifiability }\label{sec:appendix-identifiability}

This appendix gives the full protocol and results behind the identifiability claim in \S\ref{sec:irt-limitations}, via a simulation study (parameter recovery from a known ground truth) and a seed-stability test (consistency of the fit on real data across SVI runs).

\begin{figure}[h!]
    \centering
    \includegraphics[width=0.85\linewidth]{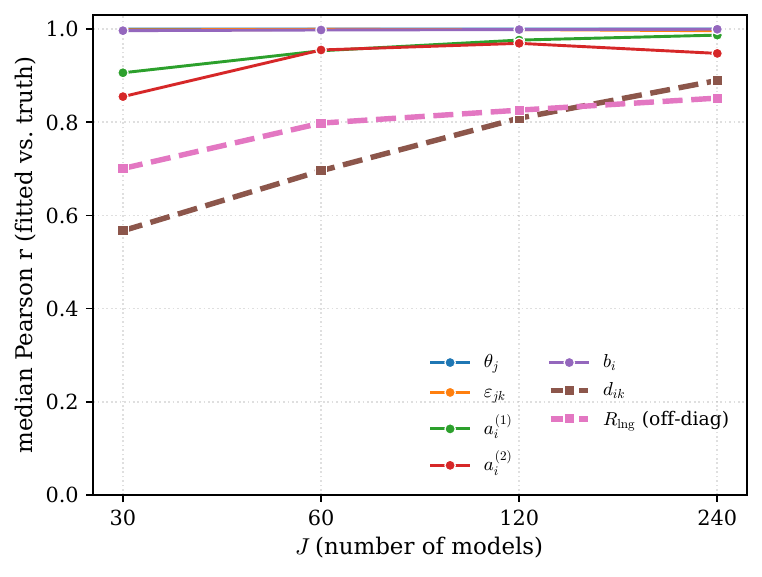}
    \caption{Parameter recovery (median Pearson $r$) versus $J$, the number of LLMs in the simulation. }
    \label{fig:identifiability-scaling}
\end{figure}

\paragraph{Simulation study.}
We sample ground-truth parameters $(\thetaj, \epsj, \aione, \aitwo, \bi, \dil, \siglng, \Rlng)$ from the prior assumed by \multiirt{} (hyperparameters in Appendix~\ref{sec:reparam}), generate responses $\Yijl \sim \Bernoulli{\pijl}$ at the real benchmark's $(I, J, L) = (\numitems, \nummodels, \numlangs)$ dimensions, refit \multiirt{} on $Y$, and correlate the fitted posterior means against the ground truth. We repeat this across $S = \numseeds$ independent draws of the ground truth and report median Pearson $r$ per parameter family (Table~\ref{tab:sim-recovery}). We additionally re-run the simulation at $J \in \{30, 60, 120, 240\}$ and report recovery as a function of $J$ (Figure~\ref{fig:identifiability-scaling}).

\paragraph{Seed stability.}
We refit \multiirt{} on the full \mmluprox{} response tensor for $S = 10$ independent SVI seeds, holding data and priors fixed, and compute the Pearson correlation between fitted posterior means for every pair of seeds, separately for each parameter family. With $S = 10$ seeds this yields $\binom{10}{2} = 45$ pairwise correlations per parameter; we report the median, minimum, and maximum across pairs (Table~\ref{tab:seed-stability}). For $\Rlng$ we correlate the off-diagonal entries of $\Lcorr \Lcorr^\top$; for the scalar $\siglng$, where Pearson is undefined, we report the relative spread $(\max - \min) / \mathrm{mean}$ across seeds.

\paragraph{Effect of the prior.}
We refit \multiirt{} on the full \mmluprox{} response tensor under four alternative prior regimes: \textit{weak} (roughly $2\times$ the manuscript scales, $\mathrm{LKJ}(1.5)$); \textit{strong} (roughly half the manuscript scales, $\mathrm{LKJ}(4)$); \textit{sparse} ($\dil \sim \mathrm{Laplace}(0, 0.15)$, other scales unchanged); and \textit{sparse hier.} ($\dil \sim \mathrm{Laplace}(0, \lambda)$ with $\lambda \sim \mathrm{HalfNormal}(0.15)$). For each regime we compute the Pearson correlation between its posterior mean and the manuscript posterior mean, per parameter family; $\siglng$ is a scalar, so we report the ratio to the manuscript value instead (Table~\ref{tab:prior-sensitivity}).

\input{tables/sim_recovery_table}

\input{tables/seed_stability_table}

\input{tables/prior_sensitivity_table}

\input{figures/translation_error_examples}

\input{figures/cultural_specific_examples}

%% file: tables/models_table.tex
\begin{table}[h]
    \centering
    \small
    \setlength{\tabcolsep}{4pt}
    \resizebox{\columnwidth}{!}{%
    \begin{tabular}{l l l l}
        \toprule
        Family & Model & Params & Variant \\
        \midrule
        \multirow{3}{*}{\aya}
            & {tiny-aya-global}       & 3B     & Instruct \\
            & {aya-expanse-8b}        & 8B     & Instruct \\
            & {aya-expanse-32b}       & 32B    & Instruct \\
        \midrule
        \multirow{6}{*}{\qwen}
            & {Qwen3.5-2B}            & 2B     & Instruct (dense) \\
            & {Qwen3.5-4B}            & 4B     & Instruct (dense) \\
            & {Qwen3.5-9B}            & 9B     & Instruct (dense) \\
            & {Qwen3.5-27B}           & 27B    & Instruct (dense) \\
            & {Qwen3.5-35B-A3B}       & 35B / 3B act.   & MoE \\
            & {Qwen3.5-122B-A10B}     & 122B / 10B act. & MoE \\
        \midrule
        \multirow{4}{*}{\olmo}
            & {Olmo-3-7B-Instruct}    & 7B     & Instruct \\
            & {Olmo-3-7B-Think}       & 7B     & Reasoning \\
            & {Olmo-3.1-32B-Instruct} & 32B    & Instruct \\
            & {Olmo-3.1-32B-Think}    & 32B    & Reasoning \\
        \midrule
        \multirow{3}{*}{\gemma}
            & {gemma-3-4b-it}         & 4B     & Instruct \\
            & {gemma-3-12b-it}        & 12B    & Instruct \\
            & {gemma-3-27b-it}        & 27B    & Instruct \\
        \midrule
        \multirow{3}{*}{\llama}
            & {Llama-3.2-3B-Instruct} & 3B     & Instruct \\
            & {Llama-3.1-8B-Instruct} & 8B     & Instruct \\
            & {Llama-3.3-70B-Instruct}& 70B    & Instruct \\
        \midrule
        \multirow{4}{*}{\phimodel}
            & {Phi-4-mini-instruct}   & 3.8B   & Instruct \\
            & {Phi-4-mini-reasoning}  & 3.8B   & Reasoning \\
            & {Phi-4-reasoning}       & 14B    & Reasoning \\
            & {Phi-4-reasoning-plus}  & 14B    & Reasoning (RL) \\
        \midrule
        \multirow{2}{*}{\gpt}
            & {gpt-oss-20b}           & 21B / 3.6B act.  & MoE, Reasoning \\
            & {gpt-oss-120b}          & 117B / 5.1B act. & MoE, Reasoning \\
        \bottomrule
    \end{tabular}}
    \caption{The \nummodels{} LLMs evaluated in this work, grouped by family and ordered by size within each family.
    }
    \label{tab:appendix-models}
\end{table}

%% file: tables/languages_table.tex
\begin{table}[h!]
    \centering
    \small
    \setlength{\tabcolsep}{6pt}
    \resizebox{\columnwidth}{!}{%
    \begin{tabular}{l l @{\hspace{2em}} l l}
        \toprule
        Code & Language & Code & Language \\
        \midrule
        af & Afrikaans   & ne & Nepali \\
        ar & Arabic      & pt & Portuguese \\
        bn & Bengali     & ru & Russian \\
        cs & Czech       & sr & Serbian \\
        de & German      & sw & Swahili \\
        en & English     & te & Telugu \\
        es & Spanish     & th & Thai \\
        fr & French      & uk & Ukrainian \\
        hi & Hindi       & ur & Urdu \\
        hu & Hungarian   & vi & Vietnamese \\
        id & Indonesian  & wo & Wolof \\
        it & Italian     & yo & Yoruba \\
        ja & Japanese    & zh & Chinese (Simpl.) \\
        ko & Korean      & zu & Zulu \\
        mr & Marathi     &    &  \\
        \bottomrule
    \end{tabular}}
    \caption{The \numlangs{} languages evaluated in this work, with their ISO 639-1 codes.
    }
    \label{tab:appendix-langs}
\end{table}

%% file: figures/judge_prompt.tex
\begin{lstlisting}[style=promptstyle, breakindent=1pt, caption={Prompt template for the translation-error judge (\S\ref{sec:translation-errors}).}, label={lst:judge-prompt}]
You are an expert multilingual linguist and an objective translation quality judge.
Your task is to evaluate the quality of a translated multiple-choice question from English to {target_language}. You must determine if the translation accurately preserves the academic meaning, logic, and formatting of the source text.

You will be provided with the English source text and the Target translation.

Evaluate the translation against the following Error Taxonomy:
1. Critical Semantic Shift: Does the translation reverse or fundamentally change the core meaning? (e.g., mistranslating "buying" as "selling", or "sphere" as "sky").
2. Logic Alteration: Does the translation alter the fundamental truth value of the options? (e.g., accidentally translating an explicitly false multiple-choice option in a way that makes it factually true).
3. Source Intrusion: Are English source words lazily left in parentheses alongside the translation instead of relying on proper localized terminology?
4. Formatting & Localization Failure: Are numbers, decimal separators, LaTeX formulas, or typographic symbols poorly adapted to the target locale conventions?

Provide your evaluation as a valid JSON object using the exact schema below. Do not output any markdown formatting, conversational text, or explanations outside of the JSON structure.

JSON Schema:
{
  "has_error": true/false,
  "severity": "Critical" | "Minor" | "None",
  "primary_error_category": "Semantic Shift" | "Logic Alteration" | "Source Intrusion" | "Formatting Failure" | "None",
  "explanation": "A concise, 1-2 sentence explanation of the specific error and the exact mistranslated word/phrase. If no error, output 'Accurate translation.'"
}

--- SOURCE (English) ---
Question:
{en_question}

Options:
{en_options}

Correct answer: {en_answer_letter}) {en_answer_text}

--- TRANSLATION ({target_language}) ---
Question:
{tgt_question}

Options:
{tgt_options}

Correct answer: {tgt_answer_letter}) {tgt_answer_text}
\end{lstlisting}

%% file: figures/cultural_judge_prompt.tex
\begin{lstlisting}[style=promptstyle, breakindent=1pt, caption={Prompt template sent to \gemini{} for the cultural-specificity judge (\S\ref{sec:culture}.}, label={lst:cult-judge-prompt}]
You are an expert evaluator of multiple-choice exam questions.

Your task is to determine whether answering the following question requires *culture-specific knowledge* (e.g., knowledge tied to a particular country, region, legal system, religion, language community, named entity, social custom, regional cuisine, currency, or local convention).

A question is UNIVERSAL if its content is broadly applicable across cultures: math, hard-science facts, general logic, biology, chemistry, physics, generic programming, and similar.

A question is CULTURALLY SPECIFIC if it tests knowledge that:
- references a particular country's laws, history, geography, politics, or institutions ("U.S. tax law", "the Indian Constitution", "the EU GDPR"),
- references a particular religious or philosophical tradition's tenets,
- requires understanding of a specific language's grammar, idiom, or wordplay beyond the surface meaning,
- references region-specific named entities (specific authors, cuisines, athletes, brands, or local conventions),
- assumes social conventions or value systems tied to a particular culture.

Provide your evaluation as a valid JSON object using the exact schema below.
Do not output any markdown formatting, conversational text, or explanations
outside of the JSON structure.

JSON Schema:
{
  "is_culturally_specific": true/false,
  "specificity_type": "Universal" | "Region/Country" | "Religion/Philosophy" | "Language-internal" | "Named-entity" | "Social-convention" | "None",
  "region": "ISO code, country, or 'None' if not applicable",
  "confidence": "high" | "medium" | "low",
  "explanation": "A concise, 1-2 sentence explanation. If universal, output 'Universal: <subject area>.'"
}

--- QUESTION (English source) ---
Subject: {category}

Question:
{question}

Options:
{options}

Correct answer: {answer_letter}) {answer_text}
\end{lstlisting}

%% file: figures/translation_instructions.tex
\begin{figure*}[ht!]
\centering
\begin{tcolorbox}[
    colback=gray!5,
    colframe=gray!55,
    boxrule=0.4pt,
    arc=3pt,
    left=10pt, right=10pt, top=6pt, bottom=6pt,
    title={\textsc{Annotator instructions: Translation errors}\hfill\itshape\footnotesize (human-validation app)},
    coltitle=white,
    colbacktitle=gray!60,
    fonttitle=\bfseries\footnotesize,
    enhanced jigsaw,
]
\footnotesize
\textbf{Task.} You will see a multiple-choice question that an AI judge flagged as having a translation error from English into your target language. Decide whether you \emph{agree} (an error exists and the judge's explanation correctly identifies it), \emph{partially agree} (an error exists but the explanation is wrong, incomplete, or misses its full scope; a short note is required), or \emph{disagree} (no real error; meaning and formatting are preserved).

\smallskip
\textbf{Translation error categories:}
\begin{itemize}\setlength{\itemsep}{0pt}\setlength{\parskip}{0pt}
  \item \textbf{Semantic Shift}: the translation changes the meaning of a word, phrase, or option (e.g., ``buying'' rendered as ``selling'').
  \item \textbf{Logic Alteration}: the truth value of an answer option changes (e.g., a false option becomes factually true).
  \item \textbf{Source Intrusion}: English source words are left in parentheses instead of proper target-language terminology.
  \item \textbf{Formatting Failure}: numbers, decimal separators, LaTeX, or symbols are not adapted to the target locale, or options are duplicated/shuffled.
\end{itemize}
\end{tcolorbox}
\caption{Instructions shown to annotators for the translation-error judge, defining the agree/partially-agree/disagree labels and the four error categories. Reading tips omitted for space.}
\label{lst:translation-instructions}
\end{figure*}

%% file: figures/validation_instructions.tex
\begin{figure*}[t]
\centering
\begin{tcolorbox}[
    colback=gray!5,
    colframe=gray!55,
    boxrule=0.4pt,
    arc=3pt,
    left=10pt, right=10pt, top=6pt, bottom=6pt,
    title={\textsc{Annotator instructions}\hfill\itshape\footnotesize (cultural-specificity judge)},
    coltitle=white,
    colbacktitle=gray!60,
    fonttitle=\bfseries\footnotesize,
    enhanced jigsaw,
]
\footnotesize
\textbf{Task.} Decide whether a multiple-choice question requires knowledge tied to a specific culture, region, religion, or language, or whether the answer is universal across cultures.

\smallskip
\textbf{Culturally Specific} if the question references a specific country's laws/history/geography/institutions; a specific religious or philosophical tradition's tenets; a specific language's grammar, idiom, or wordplay; region-specific named entities (authors, cuisines, brands, conventions); or social conventions tied to a particular culture.

\smallskip
\textbf{Universal} if the content holds across cultures: math, hard sciences, generic programming, and similar. A hard-science or programming question that merely \emph{mentions} a country or language is still Universal.

\smallskip
\textbf{Specificity types} (if culturally specific):
\begin{itemize}\setlength{\itemsep}{0pt}\setlength{\parskip}{0pt}
  \item \textbf{Region/Country}: a specific country's laws, history, or institutions.
  \item \textbf{Religion/Philosophy}: a religious or philosophical tradition.
  \item \textbf{Language-internal}: a specific language's grammar or wordplay.
  \item \textbf{Named-entity}: a culturally-embedded person, work, or brand.
  \item \textbf{Social-convention}: social customs or value systems.
  \item \textbf{Other}: culturally specific but none of the above.
\end{itemize}
\end{tcolorbox}
\caption{Instructions shown to annotators for the cultural-specificity judge, defining Culturally Specific vs.\ Universal and the specificity sub-types. Worked examples and edge-case guidance omitted for space.}
\label{lst:validation-instructions}
\end{figure*}

%% file: figures/translation_overlap_mgsm.tex
\begin{figure}[ht!]
\centering
\includegraphics[width=0.9\linewidth]{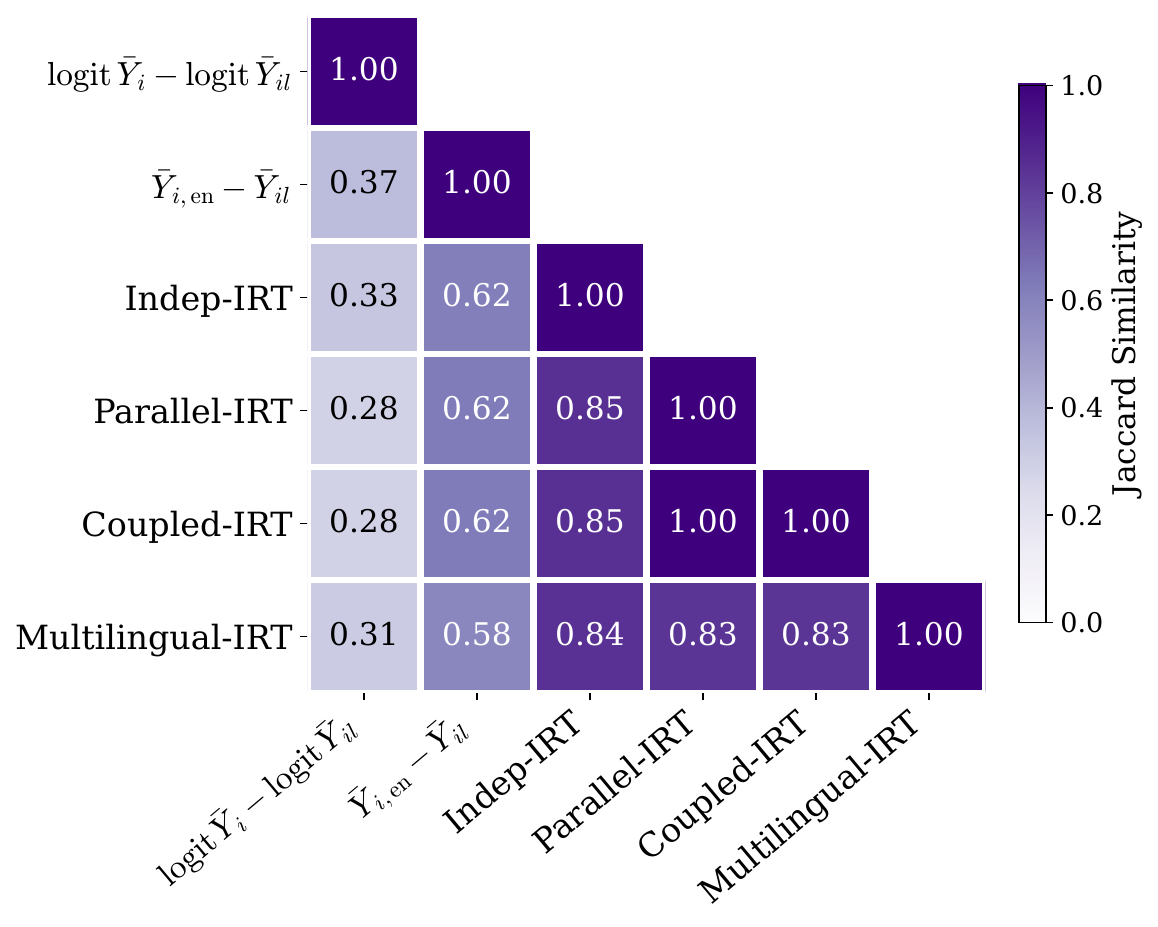}
\caption{\textbf{Top-250 critical-strict translation errors surfaced by six scoring methods on \mgsm{} (\S\ref{sec:translation-errors}).} Pairwise Jaccard overlap between top-scored translation errors. The IRT difficulty-shift score $s(i,l) {=} \dil/\widehat{\mathrm{SE}}(\dil)$ is computed for \multiirt{} ($n{=}116$), \coupledirt{} ($n{=}111$), \parallelirt{} ($n{=}111$), and \indepirt{} ($n{=}118$). The accuracy-based non-parametric baselines are an English-anchored difference $\bar Y_{i,\mathrm{en}}-\bar Y_{il}$ ($n{=}110$) and a logit-additive analog $\logit\,\bar Y_i - \logit\,\bar Y_{il}$ ($n{=}70$), both averaged across the $\nummodels$ LLMs, with $\bar Y_{i}$ additionally averaged across all languages.}
\label{fig:jaccard_mgsm}
\end{figure}

%% file: figures/cultural_jaccard.tex
\begin{figure}[h!]
    \centering
    \includegraphics[width=0.9\linewidth]{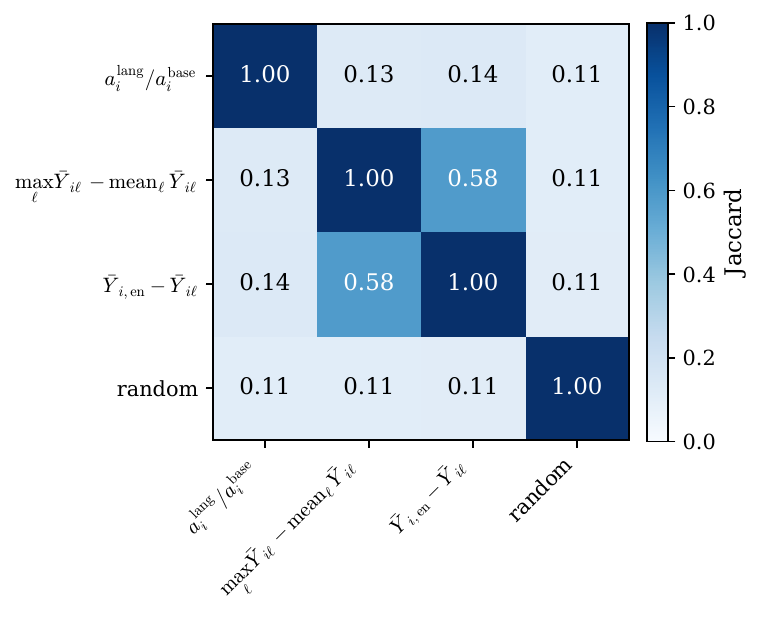}
    \caption{Pairwise Jaccard overlap between the cultural-positive sets surfaced by each ranker at top-$2000$ (\S\ref{sec:culture}).
    Accuracy-based non parametric baselines are: (1) $\max_l \bar Y_{i l}{ - }\mathrm{mean}_l \bar Y_{i l}$ (``one language stands out''), and (2) the English-anchored gap $\mathrm{mean}_{l \neq \mathrm{en}}(\bar Y_{i,\mathrm{en}} {-} \bar Y_{i l})$ (``English is consistently easier than other languages''). A uniform-random baseline is also used to mark the base rate of cultural specific items in the benchmark.
    }
    \label{fig:cultural-jaccard}
\end{figure}

%% file: tables/stability_test_for_subsets.tex
\begin{table}[t]
\centering
\resizebox{\columnwidth}{!}{%
\begin{tabular}{lcccc}
\toprule
Condition & $J$ & $r$ off-diag & mean $|\Delta R|$ & family gap \\
\midrule
\multicolumn{5}{l}{\emph{Null distributions} (range over refits)} \\
SVI seed noise ($\times 10$)       & 25 & .85--.97 & .06--.13 & .27--.29 \\
Leave-one-LLM-out ($\times 25$)   & 24 & .92--.99 & .03--.09 & .25--.30 \\
Random 3-LLM drop ($\times 10$)   & 22 & .80--.96 & .06--.15 & .25--.29 \\
\midrule
\multicolumn{5}{l}{\emph{Targeted drops}} \\
3 largest MoE                     & 22 & .86 & .12 & .30 \\
All 4 MoE                         & 21 & .91 & .10 & .33 \\
3 highest-ability                 & 22 & .83 & .14 & .28 \\
\bottomrule
\end{tabular}%
}
\caption{\textbf{Stability analysis of $\Rlng$ under different sets of LLMs of \mmluprox{}.}
Each row refits \multiirt{} on an LLM-ablated response tensor, holding priors, SVI recipe and seed fixed, and compares the fitted $\Rlng$ to the full-pool fit: $r$ off-diag and mean $|\Delta R|$ measure how far it moved, and the family gap (mean within-family minus mean between-family correlation) measures whether the linguistic structure survived. The three targeted drops fall inside the random-drop null on both distance measures, and none reduces the family gap below its range.}
\label{tab:rlng-llm-ablation}
\end{table}

%% file: tables/sim_recovery_table.tex
\begin{table}[b]
    \centering
    \resizebox{0.9\columnwidth}{!}{%
    \begin{tabular}{lccc}
        \toprule
        Parameter & median $r$ & min $r$ & max $r$ \\
        \midrule
        $\theta_j$ & 1.000 & 1.000 & 1.000 \\
        $\varepsilon_{jl}$ & 0.998 & 0.974 & 1.000 \\
        $a_i^{(1)}$ & 0.897 & 0.839 & 0.925 \\
        $a_i^{(2)}$ & 0.853 & 0.357 & 0.961 \\
        $b_i$ & 0.996 & 0.994 & 0.996 \\
        $d_{il}$ & 0.532 & 0.508 & 0.540 \\
        $R_{\text{lng}}$ (off-diag) & 0.643 & 0.614 & 0.724 \\
        \midrule
        $\sigma_{\text{lng}}$ (rel.\ err., median) & \multicolumn{3}{c}{$-24.0\%$} \\
        \bottomrule
    \end{tabular}}
    \caption{Parameter recovery in the simulation study (\S\ref{sec:appendix-identifiability}). Pearson $r$ between ground truth and fitted posterior mean across $S$ independent draws.}
    \label{tab:sim-recovery}
\end{table}

%% file: tables/seed_stability_table.tex
\begin{table}[t]
    \centering
    \resizebox{\columnwidth}{!}{%
    \begin{tabular}{lccc}
        \toprule
        Parameter & median $r$ & min $r$ & max $r$ \\
        \midrule
        $\theta_j$                          & 1.000 & 1.000 & 1.000 \\
        $\varepsilon_{jk}$                  & 1.000 & 1.000 & 1.000 \\
        $a_i^{(1)}$                         & 0.998 & 0.997 & 0.998 \\
        $a_i^{(2)}$                         & 0.996 & 0.996 & 0.997 \\
        $b_i$                               & 1.000 & 1.000 & 1.000 \\
        $d_{ik}$                            & 0.944 & 0.944 & 0.945 \\
        $R_{\text{lng}}$ (off-diag)         & 0.965 & 0.849 & 0.998 \\
        \midrule
        $\sigma_{\text{lng}}$ (rel.\ spread) & \multicolumn{3}{c}{8.916\%} \\
        \bottomrule
    \end{tabular}}
    \caption{Pairwise Pearson correlation across $S{=}10$ independent SVI seeds for the \multiirt{} fit on the full \mmluprox{} response tensor. With $S{=}10$ this is the distribution of ${\binom{10}{2}}{=}45$ pairwise correlations per parameter family.}
    \label{tab:seed-stability}
\end{table}

%% file: tables/prior_sensitivity_table.tex
\begin{table}[t]
    \centering
    \small
    \begin{tabular}{lcccc}
        \toprule
        Parameter & weak & strong & sparse & sparse hier. \\
        \midrule
    $\theta_j$                     & 1.000 & 1.000 & 1.000 & 1.000 \\
    $\varepsilon_{jl}$             & 0.999 & 0.999 & 1.000 & 0.999 \\
    $a_i^{(1)}$                    & 0.997 & 0.994 & 1.000 & 0.999 \\
    $a_i^{(2)}$                    & 0.993 & 0.994 & 0.999 & 0.995 \\
    $b_i$                          & 0.999 & 0.999 & 1.000 & 1.000 \\
    $d_{il}$                       & 0.991 & 0.950 & 0.918 & 0.932 \\
    $R_{\text{lng}}$ off-diag      & 0.998 & 0.984 & 1.000 & 0.970 \\
    $\sigma_{\text{lng}}$ (ratio)  & 0.961 & 1.024 & 0.989 & 1.016 \\
        \bottomrule
    \end{tabular}
    \caption{Pearson correlation between the manuscript posterior mean and the posterior under the alternate prior regime. The bottom row reports the ratio of fitted $\siglng$ to the manuscript value (a scalar, so Pearson is undefined).}
    \label{tab:prior-sensitivity}
\end{table}

%% file: figures/translation_error_examples.tex
\begin{figure*}[t]
    \centering
    \includegraphics[width=\textwidth]{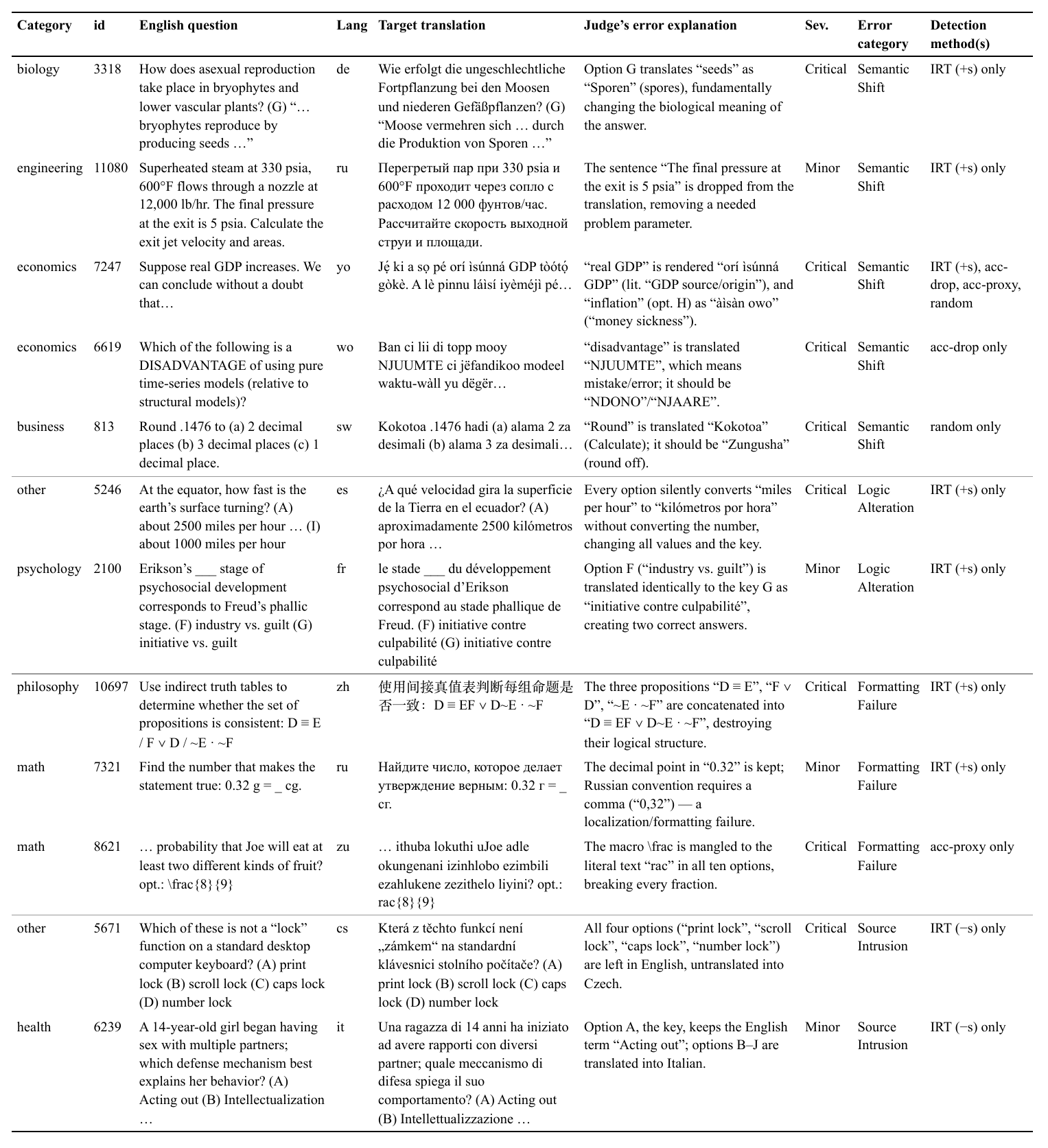}
    \caption{\gemini{} LLM-as-a-judge verdicts on machine-translated \mmluprox{} items, spanning the four error categories and both severities (\S\ref{sec:translation-errors}).}
    \label{fig:translation-error-examples}
\end{figure*}

%% file: figures/cultural_specific_examples.tex
\begin{figure*}[t]
    \centering
    \includegraphics[width=\textwidth]{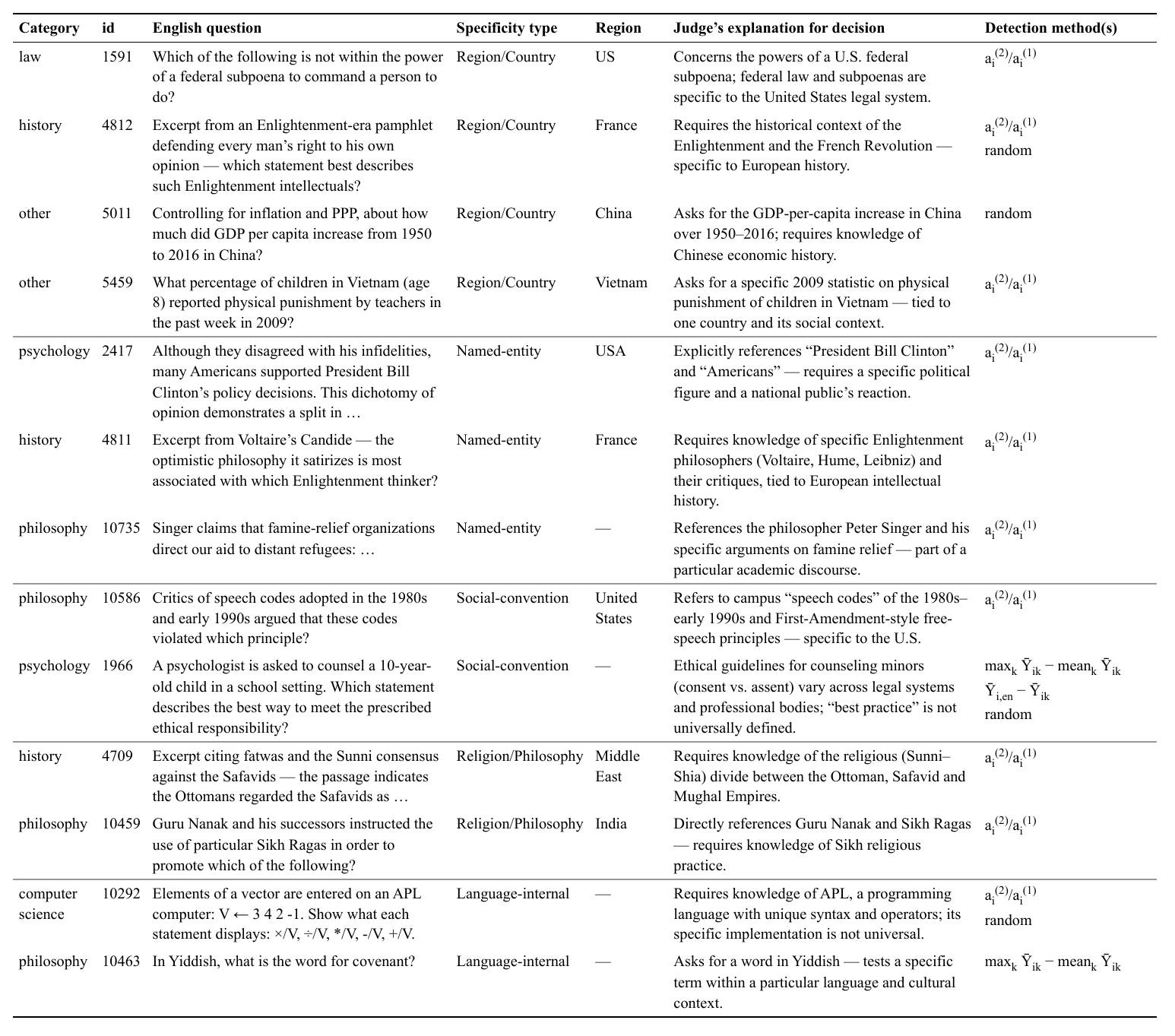}
    \caption{Real \gemini{} LLM-as-a-judge verdicts on culturally-specific \mmluprox{} items, spanning all five cultural \texttt{specificity\_type}s
    (\S\ref{sec:culture}).}
    \label{fig:cultural-specific-examples}
\end{figure*}